\documentclass[10pt]{article} 
\usepackage[accepted]{tmlr}

\usepackage{amsmath,amsfonts,bm}

\def\eqref#1{equation~\ref{#1}}

\def\1{\bm{1}}

\DeclareMathAlphabet{\mathsfit}{\encodingdefault}{\sfdefault}{m}{sl}
\SetMathAlphabet{\mathsfit}{bold}{\encodingdefault}{\sfdefault}{bx}{n}

\usepackage{xcolor}

\usepackage{amsmath,amssymb,amsthm}
\usepackage{algorithm}
\usepackage{algpseudocode}
\usepackage{array}
\usepackage{booktabs}
\usepackage{capt-of}
\usepackage{multirow}
\usepackage{wrapfig}
\usepackage{adjustbox}
\usepackage[percent]{overpic}
\usepackage{bm}

\usepackage{enumitem}

\usepackage{tabularx}
\usepackage{ragged2e}
\usepackage{url}
\usepackage{hyperref}

\usepackage{fontawesome5}

\newtheorem{assumption}{Assumption}
\newtheorem{theorem}{Theorem}

\newtheorem{proposition}{Proposition}

\renewcommand{\sectionautorefname}{Section}
\renewcommand{\subsectionautorefname}{Subsection}

\usepackage{pifont}
\newcommand{\cmark}{\textcolor{green}{\ding{51}}}%
\newcommand{\xmark}{\textcolor{red}{\ding{55}}}%

\usepackage{etoolbox}

\preto\appendix{%
  \renewcommand{\sectionautorefname}{Appendix}%
  \renewcommand{\subsectionautorefname}{\sectionautorefname}%
}

\definecolor{revisionthree}{RGB}{180,35,35}

\title{A Few Large Shifts: Layer-Inconsistency Based Minimal Overhead Adversarial Example Detection}

\author{\name Sanggeon Yun \email sanggeoy@uci.edu \\
      \addr Department of Computer Science\\
      University of California, Irvine
      \AND
      \name Ryozo Masukawa \email rmasukaw@uci.edu \\
      \addr Department of Computer Science\\ University of California, Irvine
      \AND
      \name Hyunwoo Oh \email hyunwooo@uci.edu \\
      \addr Department of Computer Science\\ University of California, Irvine
      \AND
      \name Nathaniel D. Bastian \email ndbastian@jhu.edu \\
      \addr Whiting School of Engineering\\ Johns Hopkins University
      \AND
      \name Mohsen Imani \email m.imani@uci.edu \\
      \addr Department of Computer Science\\ University of California, Irvine}

\def\month{08}  
\def\year{2026} 
\def\openreview{\url{https://openreview.net/forum?id=0xXYBxDNHA}} 

\begin{document}

\maketitle

\begin{abstract}\label{abs}
    Deep neural networks (DNNs) are highly susceptible to adversarial examples---small, malicious perturbations that can cause incorrect predictions. We introduce a lightweight, plug-in detector that uses internal layer-wise inconsistencies within the target model and requires only benign data for fitting and calibration. The approach is motivated by the \textbf{A Few Large Shifts Assumption}, an empirical hypothesis that adversarial perturbations often produce large, localized growth in representation changes across a small number of consecutive layers, connecting adversarial behavior to layer-wise Lipschitz continuity. We develop two complementary scores---\textbf{Recovery Testing (RT)} for intermediate-layer inconsistency and \textbf{Logit-layer Testing (LT)} for augmentation-induced output instability---and fuse them through RLT. Across CIFAR-10, CIFAR-100, and ImageNet, RLT achieves strong detection performance under standard attacks with substantially lower overhead than detector families requiring external encoders or reference-set retrieval. We further study its behavior under adaptive attacks, at low false-positive operating points, and under benign distribution~shifts. The code is available here: \faIcon{github}\ \url{https://github.com/c0510gy/AFLS-AED}.
    
\end{abstract}

\section{Introduction}\label{sec1}

\begin{figure}[ht]
    \centering
    \begin{overpic}[width=1.0\linewidth]{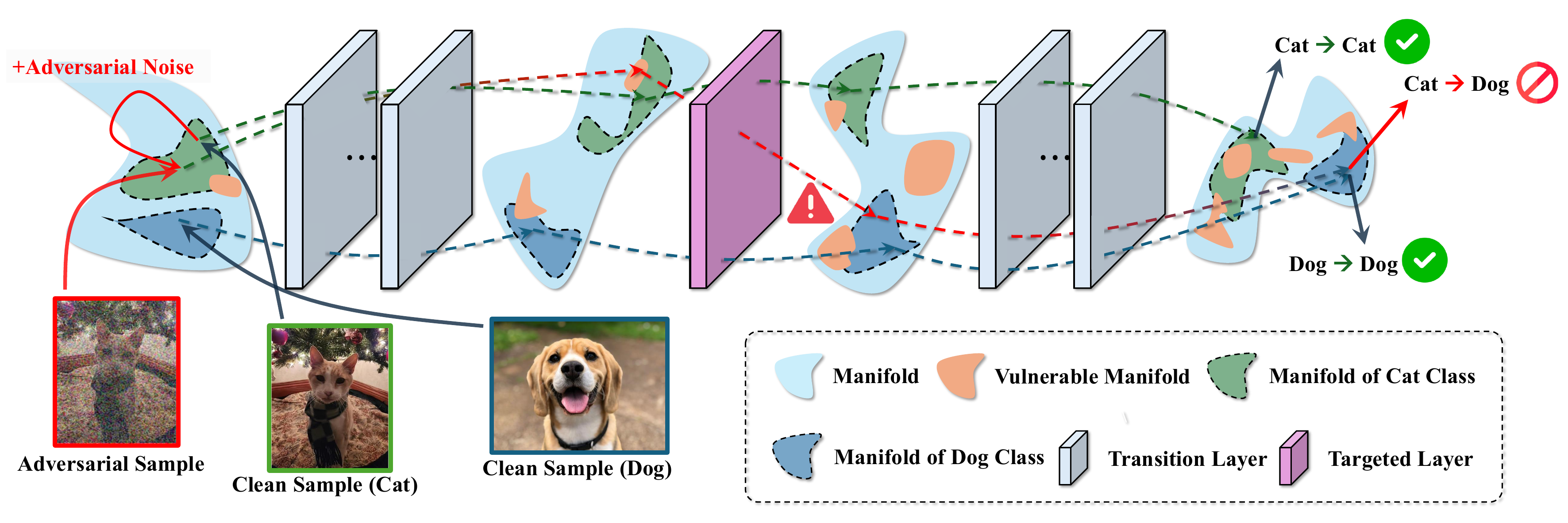}
    \end{overpic}
    \vspace{-9mm}
    \caption{\textbf{Conceptual illustration of the A Few Large Shifts hypothesis.} An adversarial perturbation moves an input across the decision boundary while producing a pronounced representation change at a small number of consecutive layers, motivating detection through layer-wise inconsistency.}
    \label{fig:manifold_figure_v2}
\end{figure}
Deep neural networks (DNNs) have been broadly deployed in computer vision, natural language processing, multi-modal tasks, and beyond~\cite{nan2024di, khachatryan2023text2video, zhang2024slicing}. Although they exhibit remarkable performance, \emph{adversarial examples (AEs)}---inputs containing small yet malicious perturbations---can induce misclassifications while appearing virtually unchanged to human observers~\cite{demontis2019adversarial}. This vulnerability poses severe risks in high-stakes domains such as autonomous driving or disease diagnosis~\cite{zheng2024physical, ravikumar2024securing}, underlining the need for robust defenses.

Existing defensive strategies generally fall into three categories: adversarial training, input purification, and AE detection~\cite{he2022your, abusnaina2021adversarial}. 
\emph{Adversarial training} retrains or fine-tunes a model on adversarially perturbed samples to obtain an adversarially trained classifier (ATC)~\cite{wang2024revisiting, liu2024perturbation}. Although it is among the strongest empirical defenses~\cite{zhang2019theoretically}, it incurs substantial computational cost and may reduce clean-data accuracy.
\emph{Input purification} methods~\cite{mao2021adversarial, song2024mimicdiffusion} attempt to remove adversarial noise through preprocessing (e.g., denoising), yet often fail against adaptive attacks~\cite{croce2020reliable}. 

A more flexible alternative is AE detection, which avoids the costly retraining process of ATCs by instead rejecting suspicious inputs at test time~\cite{xu2017feature, zuo2021exploiting}. Detection-based defenses offer distinct advantages, including \emph{lower implementation costs}, as they do not require exhaustive adversarial training with adversarial data, and \emph{tunable robustness}, where sensitivity can be adjusted to meet application-specific accuracy–robustness trade-offs. Furthermore, their \emph{plug-and-play} nature allows them to be integrated with existing models, including ATCs, to further enhance system-level robustness. By architecturally separating the detection mechanism from the classifier, this approach introduces an additional layer of security at the system level. AE detection techniques are often categorized into two types: those that compare inputs to a reference set (e.g., Deep k-Nearest Neighbors, DkNN~\cite{papernot2018deep}; Latent Neighborhood Graph, LNG~\cite{abusnaina2021adversarial}), and those that analyze invariants in the learned representations~\cite{jiang2020robust, chen2021exploring}.
Despite these benefits, prominent detection paradigms suffer from practical limitations. Reference-based detectors often require storing adversarial examples or constructing complex neighbor graphs, which is computationally intensive. To address this, recent work~\cite{he2024beyond} has proposed detecting consistency between augmented inputs using pre-trained Self-Supervised Learning (SSL) models. However, this introduces significant overhead from large external models---11.91$\times$ the target classifier's parameter count in our ResNet-110 comparison---and assumes the availability of high-quality, domain-specific SSL models.

To eliminate the overhead of external models, complex data structures, or heavy augmentations, we study how adversarial perturbations propagate through the target network itself. Prior work suggests that such perturbations often leave early-layer features largely intact while inducing sharp deviations in deeper layers~\cite{jiang2020robust}. This motivates our central research question: \textbf{\textit{``Can the propagation of representation changes through the target model reveal adversarial inputs?''}}
To study this question, we compare how the representations of a clean input and its perturbed version change as they pass through the network (\autoref{exp3} and \autoref{app:additional_afls_validation}). \autoref{fig:layerwise_lipschitz_violation} and \autoref{tab:compact_afls_summary} show that benign transformations generally preserve the scale of representation changes across consecutive layers, whereas adversarial perturbations often produce pronounced growth at only a few transitions. We interpret this propagation pattern through a layer-wise Lipschitz perspective and formalize it in the \textbf{A Few Large Shifts Assumption}.
The paired clean--perturbed comparison provides the empirical motivation for our detector. At inference time, Recovery Testing (RT) evaluates whether intermediate representations are consistent with recovery modules trained on benign data, while Logit-layer Testing (LT) measures output changes under learned input transformations relative to the corresponding changes in intermediate features. RLT combines the two scores after normalizing them on benign data.

The distinguishing features and requirements of our method are summarized in~\autoref{tab:comparison_ours_baselines}, where we compare RT and LT against several representative baselines. As shown, our method requires no adversarial examples, external SSL model, kNN graph, or excessive augmentation, while still requiring benign-only auxiliary fitting and calibration for RT/LT. Furthermore, while prior AE detection works often omit system-level operating-point analysis, our work addresses this gap by introducing a threshold-selection method with an explicit lower bound on overall system accuracy under the stated metric assumptions, as detailed in \autoref{app:system_level_analysis}.

Experiments on supervised CNN classifiers from several families and a supervised ViT-B/16 further show that RLT transfers across distinct target architectures under the evaluated settings (\autoref{sec:arch_eval} and \autoref{app:vit}).

Our key contributions are as follows:
\begin{itemize}[leftmargin=*, nosep]
    \vspace{-1mm}
    \item We propose a novel adversarial detection paradigm that exploits partial consistency in internal feature transformations of adversarial examples. Unlike prior methods that rely on computationally expensive external reference sets or large auxiliary models, our approach eliminates this overhead by measuring inconsistencies across a model's own layers for a single input.
    \item We directly measure how representation changes grow between consecutive layers and introduce two complementary benign-only measures, \textbf{Recovery Testing (RT)} and \textbf{Logit-layer Testing (LT)}, for intermediate-layer and output-layer inconsistency.
    \item We conduct a comprehensive evaluation under standard and adaptive attacks, across low-FPR operating points and multiple target architectures. We further analyze adaptive-objective gradients, benign calibration, and system-level threshold selection.
\end{itemize}

\begin{table}[htbp]
\centering
\caption{Comparison of method requirements. 
\textbf{RT} = Recovery Testing, 
\textbf{LT} = Logit-layer Testing. 
We also include \textbf{BEYOND}~\citep{he2024beyond},
\textbf{LID}~\citep{ma2018characterizing}, 
\textbf{\citet{mao2021adversarial}}, 
\textbf{\citet{hu2019new}}, 
\textbf{DkNN}~\citep{papernot2018deep}, 
\textbf{kNN-Def.}~\citep{dubey2019defense}. 
A \cmark\, indicates that the method satisfies the criterion, 
while a \xmark\, indicates it does not. 
}\label{tab:comparison_ours_baselines}
\begin{adjustbox}{max width=0.9\textwidth}
\begin{tabular}{@{}lcc|cccccc@{}}
\toprule
& \multicolumn{2}{c}{\textbf{Ours}} & \multicolumn{6}{c}{\textbf{Baseline Methods}} \\
\cmidrule(lr){2-3} \cmidrule(lr){4-9}
\textbf{Criterion} & \textbf{RT} & \textbf{LT} 
                   & \textbf{BEYOND}
                   & \textbf{LID}
                   & \textbf{Mao}
                   & \textbf{Hu}
                   & \textbf{DkNN}
                   & \textbf{kNN-Def.}
                   \\
\midrule

\textbf{No Adversarial Training Data?}        & \cmark & \cmark  
                                                                    & \cmark & \xmark & \xmark & \xmark & \xmark & \xmark \\

\textbf{No External Pre-trained Model?} 
                                                                    & \cmark & \cmark 
                                                                      & \xmark & \cmark & \xmark & \cmark & \cmark & \cmark \\

\textbf{No Nearest-Neighbor Retrieval?}   & \cmark & \cmark 
                                                                          & \cmark & \xmark & \cmark & \cmark & \xmark & \xmark \\

\textbf{No Large Augmentation Set?}             & \cmark & \cmark 
                                                                          & \xmark & \cmark & \cmark & \cmark & \cmark & \cmark \\

\textbf{No Auxiliary Optimization?}          & \xmark & \xmark 
                                                                          & \xmark & \xmark & \xmark & \xmark & \xmark & \xmark \\

\bottomrule
\end{tabular}
\end{adjustbox}
\vspace{1em}
\footnotesize
\begin{adjustbox}{max width=\textwidth}
\begin{tabular}{@{}l@{}}
\end{tabular}
\end{adjustbox}
\normalsize
\end{table}

\begin{figure}[h]
    \centering
    \vspace{-1mm}
    \includegraphics[width=1.\linewidth]{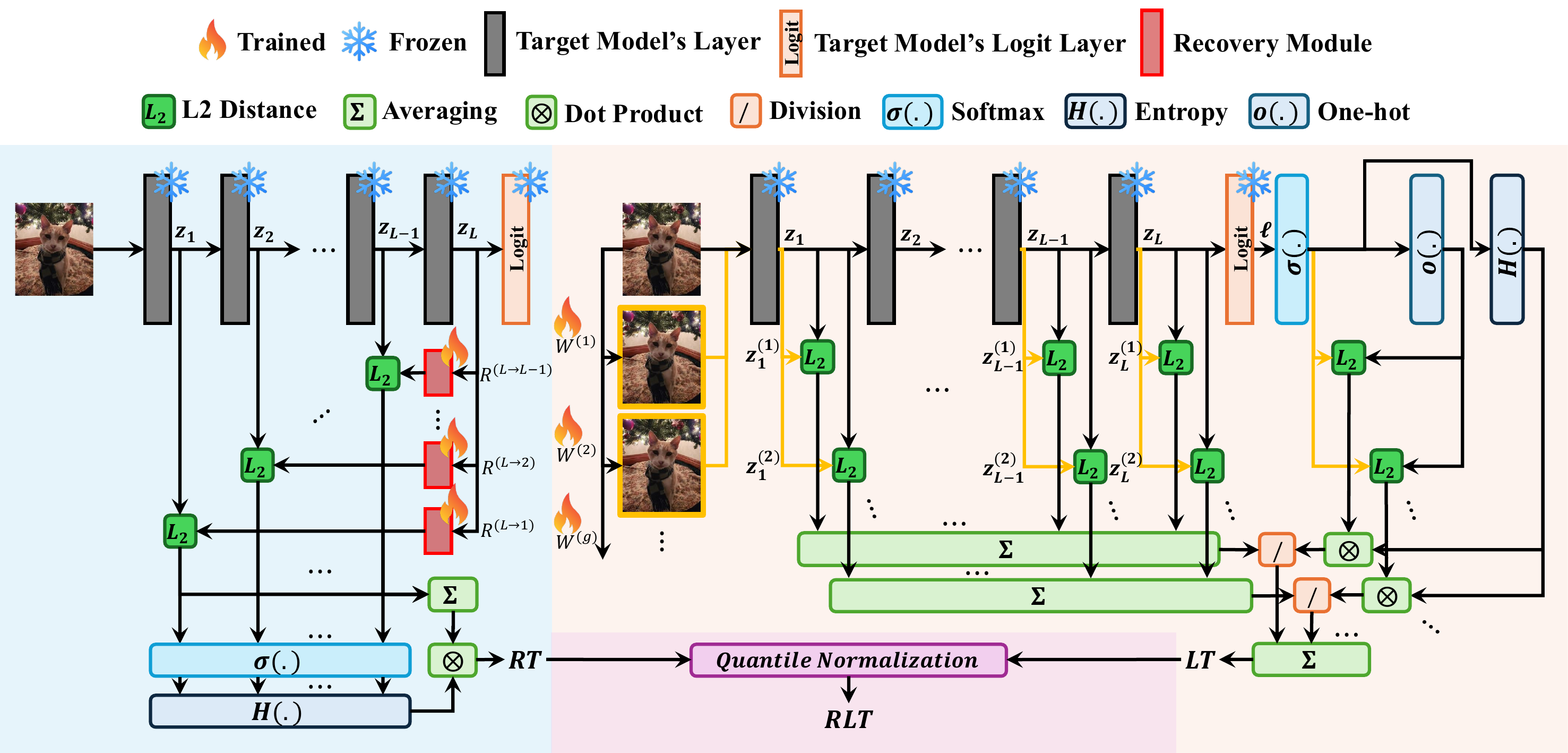}
    \vspace{-7mm}
    \caption{
        \textbf{Overview of the layer-wise detection framework.} (Left) \textit{Recovery Testing (RT)} reconstructs intermediate features $z_k$ from the final embedding $z_L$ and combines the reconstruction-error magnitude with its concentration across recovered layers. (Right) \textit{Logit-layer Testing (LT)} applies learned transformations $W^{(g)}$ and averages the output change relative to the corresponding change in intermediate features. (Bottom) \textit{Recovery and Logit Testing (RLT)} maps RT and LT through benign empirical CDFs and sums their squared normal scores, as defined in \autoref{eq:rlt}.}
    \label{fig:overview}
\vspace{-2mm}
\end{figure}
\section{Methodology}
\vspace{-2mm}

In this section, we introduce our layer-wise adversarial example detection framework (\autoref{fig:overview}), comprising three measures: RT, LT, and their fusion RLT. RT captures inconsistencies in intermediate-layer embeddings, while LT measures logit instability under augmented inputs.
At test time, we select a detection measure $m\in\{\mathrm{RT},\mathrm{LT},\mathrm{RLT}\}$ and threshold $\tau_m$. 
We classify $x$ as \emph{adversarial} when $m(x)>\tau_m$, i.e.,
$
\hat{a}(x)\;=\;\mathbb{I}\{\,m(x)>\tau_m\,\}\in\{0,1\}.
$
The benign calibration protocol is specified in \autoref{sec:calibration_protocol}; system-level operating-point selection is discussed in \autoref{app:system_level_analysis}.
The score-computation algorithm is shown in \autoref{alg:layerwise_detection}. Conditional analyses and their proofs are given in \autoref{app:assA} and \autoref{app:assB}, respectively.

\vspace{-2mm}
\subsection{Definitions and Notation}
\vspace{-2mm}

Given a target DNN $f(\cdot) = f_{logit}\circ f_L\circ f_{L-1}\circ\cdots\circ f_1(\cdot)$ with $L$ intermediate layers $f_i$ and one logit layer $f_{logit}$ at the end, we compute intermediate representations $z_i(x_j)=f_i\circ\cdots\circ f_1(x_j)\in \mathbb{R}^{D_i}$ for all $i \in \{1, 2, \cdots, L\}$, where $D_i$ denotes the dimensionality of the embedding space and $x_j$ denotes an input sample. We assume the logit layer takes the last intermediate embedding $z_L$ to produce the final logit output $\ell = f_{logit}(z_L)\in \mathbb{R}^C$, where $C$ is the number of output classes.
We denote the softmax function as $\sigma(x)=\frac{\exp{x}}{\mathbf 1^\top \exp{x}}$, the Shannon entropy as $\mathcal{H}(p) = -\sum_i p_i\log(p_i+\varepsilon)$ with a small numerical constant $\varepsilon>0$, and the class-$c$ one-hot basis vector as $\mathbf{o}_c\in\{0,1\}^C$. Labels are not used to train the auxiliary modules: all recovery modules, transformations, and thresholds are fitted/calibrated using only benign samples $\mathcal{D}_{norm}=\{x_1, x_2, \cdots, x_N\}$. At test time, LT uses the model-predicted class $\hat y=\arg\max_c \sigma_c(\ell(x))$, not the ground-truth label.

\subsection{Motivation and A Few Large Shifts Assumption}
\label{sec:afls_hypothesis}

\paragraph{Empirical motivation.}
\autoref{fig:error_distribution} shows that benign inputs tend to have comparable RT reconstruction errors across recovered layers, whereas several attacks produce one or a few much larger errors. To examine whether this pattern reflects how a perturbation propagates through the target model, we compare each clean input with a version produced by either a benign transformation or an adversarial attack and measure the representation change at every layer. Across 12 dataset/model combinations, attack-generated pairs much more frequently exhibit an amplification beyond the benign reference, and their largest amplification often accounts for most of the total amplification across layer transitions (\autoref{fig:layerwise_lipschitz_violation} and \autoref{tab:compact_afls_summary}). These observations suggest that adversarial effects can become concentrated at a small number of layer transitions rather than growing uniformly throughout the network.

\paragraph{Theoretical motivation.}
We introduce a \emph{layer-wise Lipschitz} interpretation of this empirical pattern. Instead of describing the network with a single input--output Lipschitz constant, our framing examines how a perturbation is amplified by each transformation $z_{k+1}=f_{k+1}(z_k)$. For representations $z_k(x)$ and $z_k(x')$ contained in a neighborhood $\mathcal N$, every Lipschitz constant $L_{k+1}(\mathcal N)$ valid on that neighborhood satisfies
\begin{equation}
\frac{\|z_{k+1}(x')-z_{k+1}(x)\|_2}{\|z_k(x')-z_k(x)\|_2}
\leq L_{k+1}(\mathcal N)
\end{equation}
whenever the denominator is nonzero. After correcting for representation dimensionality, a pronounced increase between consecutive layers therefore gives a large observed lower bound on the corresponding local Lipschitz constant. We interpret an amplification that exceeds the scale observed under benign transformations as a layer-wise Lipschitz violation relative to the benign reference. Under this interpretation, the observations above indicate that such violations are both substantially more frequent for adversarial perturbations and concentrated at only a few layer transitions. Motivated by this connection, we formulate the A Few Large Shifts Assumption below. \autoref{app:lipschitz} gives the detailed relation.

\begin{assumption}[\textbf{A Few Large Shifts}]\label{assump:few_shifts}
Let $f=f_{\mathrm{logit}}\circ f_L\circ\cdots\circ f_1$ be a target network. Define $z_0(x)=x$, $z_k(x)=f_k\circ\cdots\circ f_1(x)$ for $k=1,\ldots,L$, and $z_{L+1}(x)=\ell(x)=f_{\mathrm{logit}}(z_L(x))$. Let $D_k$ be the dimension of $z_k$, and let $x^{\mathrm{adv}}=x+\delta$ be an adversarially perturbed input. For $k=0,\ldots,L$, define the dimension-normalized representation change and its amplification across consecutive layers as

\begin{equation}
d_k(x,x^{\mathrm{adv}})=
\frac{\|z_k(x^{\mathrm{adv}})-z_k(x)\|_2^2}{D_k},
\qquad
A_k(x,x^{\mathrm{adv}})=
\frac{d_{k+1}(x,x^{\mathrm{adv}})}
{d_k(x,x^{\mathrm{adv}})+\varepsilon_A},
\end{equation}

where $\varepsilon_A>0$. For each transition, let $b_k>0$ be a reference bound estimated from benign transformations before evaluating attacks. For a small integer $s\ll L+1$, let $\mathcal T_s$ contain the indices of the $s$ largest values among $\{A_k(x,x^{\mathrm{adv}})\}_{k=0}^{L}$. We assume that at least one of these transitions exceeds its benign reference scale and that the selected transitions account for a substantial fraction of the total amplification:

\begin{equation}
\max_{k\in\mathcal T_s}
\frac{A_k(x,x^{\mathrm{adv}})}{b_k}>1,
\qquad
C_s(x,x^{\mathrm{adv}})=
\frac{\sum_{k\in\mathcal T_s}A_k(x,x^{\mathrm{adv}})}
{\sum_{k=0}^{L}A_k(x,x^{\mathrm{adv}})+\varepsilon_C}
\geq\rho_s,
\end{equation}

where $\varepsilon_C>0$ and $\rho_s\in(0,1]$. Thus, the representation change grows beyond its benign scale at a small number of transitions, while the remaining transitions contribute a limited fraction of the total amplification.
\end{assumption}
\vspace{-3mm}

\paragraph{Empirical evaluation of AFLS.}
We evaluate the two defining properties using the direct clean/perturbed comparisons in \autoref{exp3} and \autoref{app:additional_afls_validation}. For any paired inputs $(x,x')$, let
\begin{equation}
V(x,x')=\mathbb I\!\left\{
\max_{0\leq k\leq L}\frac{A_k(x,x')}{b_k}>1
\right\}.
\end{equation}
Here, $V=1$ indicates that at least one transition exceeds its benign reference bound. \autoref{fig:layerwise_lipschitz_violation} shows the exceedance frequency at each transition, while \autoref{tab:compact_afls_summary} reports the aggregate $\Pr(V=1)$ and mean concentration $\mathbb E[C_1]$. Across the 12 dataset/model combinations, mean $\Pr(V=1)$ is $0.0357$ for benign transformations and $0.8751$ for attack-generated inputs, a difference of $0.8394$, and the mean attack concentration is $\mathbb E[C_1]=0.7017$. These paired-input statistics support AFLS, while RT and LT translate the same layer-wise motivation into single-input scores fitted on benign data.

\subsection{Recovery Testing Measure}

For each hidden layer $k \in \{\,k_{RT},\,k_{RT}+1,\,\ldots,\,L-1\}$, we train an approximate inverse function:
$R^{(L\rightarrow k)}: \mathbb{R}^{D_L} \longrightarrow \mathbb{R}^{D_k}$
to reconstruct $z_k$ from $z_L$ with hyperparameter $k_{RT}$. Each $R^{(L\rightarrow k)}$ is implemented as a lightweight MLP with 3–4 layers and trained by a mean squared error loss:

\vspace{-2mm}
\begin{equation}
\mathcal{L}_{RT}
= \frac{1}{N} \sum_{n=1}^{N}
\sum_{k=k_{RT}}^{L-1}
\|\,z_k(x_n) - R^{(L\rightarrow k)}\bigl(z_L(x_n)\bigr)\|_2^2.
\end{equation}
\vspace{-2mm}

At test time, we define the squared reconstruction error only for the layers for which a recovery module is trained, $k\in\{k_{RT},\ldots,L-1\}$, as

\vspace{-2mm}
\begin{equation}
\begin{aligned}
e_k(x) = \|\,z_k(x) - R^{(L\rightarrow k)}(z_L(x))\|_2^2, \quad k=k_{RT},\ldots,L-1.
\end{aligned}
\end{equation}
\vspace{-2mm}

Collecting the vector $e_{RT}(x) = (e_{k_{RT}}(x),\ldots,e_{L-1}(x))$, we normalize it with softmax to obtain a distribution $\sigma(e_{RT}(x))$ across the recovered layers. The non‐uniformity of this distribution is captured using inverse entropy $\log(L-k_{RT}) - \mathcal{H}(\sigma(e_{RT}(x)))$. The final score is computed by taking the average of the raw reconstruction errors and weighting it by this information‐based term:

\vspace{-2mm}
\begin{equation}
\begin{aligned}
RT(x)
= \bigl(\log(L-k_{RT}) - \mathcal{H}(\sigma(e_{RT}(x)))\bigr) \log\Bigl(\frac{1}{L-k_{RT}}
         \sum_{k=k_{RT}}^{L-1} e_k(x)\Bigr).
\label{eq:RT}
\end{aligned}
\end{equation}
\vspace{-2mm}

We apply a logarithm to the average raw error for numerical stability. Under the sufficient conditions of \autoref{thm:rec}, an adversarial change at an intermediate layer increases the corresponding reconstruction error when the recovery module cannot reproduce that change from the final embedding. RT does not compare a query with its clean counterpart; it evaluates the query using recovery modules trained on benign data. Intuitively, the first term of \autoref{eq:RT} is larger when the reconstruction errors are concentrated at a small number of layers, while the second term measures their average magnitude.

\subsection{Logit-layer Testing Measure}

Although RT effectively detects perturbations in intermediate layers $f_i$, it cannot be applied to the final logit layer $f_{\text{logit}}$, since there is no subsequent representation from which to reconstruct the logits. To address this—and motivated by the intuition that adversarial examples aim to induce misclassification—we introduce Logit‐layer Testing (LT): we quantify uncertainty at $f_{\text{logit}}$, relative to changes in intermediate features, using data‐driven, low‐cost input augmentations.

Let $\{W^{(g)}\}_{g=1}^{G}$ be a small set (e.g., $1 \leq G \leq 6$) of image transformation matrices. Each matrix $W^{(g)}$ is initialized as an identity transformation and is then fine-tuned on benign data using gradient descent. The LT score itself, as defined in \autoref{eq:LT}, serves as the loss function for this process, encouraging the transformations to find augmentations that preserve logit stability for benign inputs, i.e., $\sigma(f(x)) \approx \sigma(f(W^{(g)}x))$, while still perturbing intermediate features. This establishes a consistent baseline against which adversarial instability can be measured.

Given a test input, for each augmentation $g$, we measure two changes between the original and augmented inputs: the intermediate-feature change $\Delta z^{(g)}(x|W^{(g)})$ and the prediction deviation $\Delta \ell^{(g)}(x|W^{(g)})$. The intermediate-feature change is measured by averaging the $L_2$ distances across the selected intermediate outputs: $\Delta z^{(g)}(x|W^{(g)}) = \frac{1}{L - k_{LT} + 1}\sum_{i=k_{LT}}^L\|z_i(x) - z_i(W^{(g)}x)\|_2^2$ with hyperparameter $k_{LT}$. The prediction deviation is the $L_2$ distance between the one-hot vector of the original predicted class and the augmented-input softmax output: $\Delta \ell^{(g)}(x|W^{(g)}) = \|\mathbf{o}_{\hat{y}} - \sigma(\ell(W^{(g)}x))\|_2^2$, where $\hat{y} = \arg\max_{c}\sigma_c(\ell(x))$.
We then combine the two quantities into $\frac{\Delta \ell^{(g)}(x|W^{(g)})}{\Delta z^{(g)}(x|W^{(g)})}$, weight it by the entropy of the logit score vector $\mathcal{H}(\sigma(\ell(x)))$ to emphasize inputs closer to the decision boundary~\cite{galil2021disrupting}, and finally average over all augmentations as follows:

\vspace{-2mm}
\begin{equation}
LT(x) = \frac{1}{G}\sum_{g=1}^G{\log{\left(\mathcal{H}(\sigma(\ell(x)))\Delta \ell^{(g)}(x|W^{(g)})+\varepsilon\right)} - \log{\left(\Delta z^{(g)}(x|W^{(g)})+\varepsilon\right)}}.
\label{eq:LT}
\end{equation}
\vspace{-2mm}

As before, the logarithm and $\varepsilon$ ensure numerical stability. LT increases when the prediction deviation under the learned transformations is unusually large relative to the corresponding change in intermediate features. \autoref{thm:logit_amp} characterizes this behavior under an explicit output-instability condition.

To fine-tune the transformation matrices $\{W^{(g)}\}$, we directly use $LT(x)$ as the training loss:

\vspace{-2mm}
\begin{equation}
\mathcal L_{LT} = \frac{1}{N}\sum_{n=1}^{N} LT(x_n).
\end{equation}
\vspace{-2mm}

\subsection{Recovery and Logit Testing Combined Measure}

To capture inconsistencies in both intermediate and logit layers, we introduce a combined score called Recovery and Logit Testing. This score integrates RT and LT while correcting for differences in their statistical distributions using quantile normalization. Specifically, we transform each score based on its empirical cumulative distribution estimated from the benign training set, and map it to the standard normal distribution.

Let $\hat{\mathcal{F}}_{RT}$ and $\hat{\mathcal{F}}_{LT}$ be empirical CDFs fitted on $n_{\mathrm{cdf}}$ benign samples that are independent of final threshold calibration. To avoid infinite normal scores, for $m\in\{RT,LT\}$ we use the clipped mid-rank transform

\begin{equation}
\widetilde{\mathcal F}_{m}(s)=\operatorname{clip}\!\left(
\frac{\sum_{i=1}^{n_{\mathrm{cdf}}}\mathbb I\{m_i\leq s\}+1/2}{n_{\mathrm{cdf}}},
\frac{1}{2n_{\mathrm{cdf}}},1-\frac{1}{2n_{\mathrm{cdf}}}\right).
\end{equation}

Let $\Phi^{-1}$ denote the standard-normal quantile function. At test time,

\vspace{-2mm}
\begin{equation}
RT_{norm}(x) = \Phi^{-1} \left( \widetilde{\mathcal{F}}_{RT}(RT(x)) \right), \quad
LT_{norm}(x) = \Phi^{-1} \left( \widetilde{\mathcal{F}}_{LT}(LT(x)) \right).
\end{equation}
\vspace{-2mm}
The final RLT score is computed by summing the squared normalized values:
\begin{equation}
RLT(x) = (RT_{norm}(x))^2 + (LT_{norm}(x))^2.
\label{eq:rlt}
\end{equation}
\vspace{-5mm}

This transformation places both scores on a common marginal scale while retaining their dependence. The squared sum assigns a large RLT value when either normalized component lies far in a benign tail. \autoref{thm:rlt} characterizes separation under an explicit adversarial score-margin condition.

\subsection{Benign Calibration and Deployment Protocol}\label{sec:calibration_protocol}
We separate benign data into three roles: (i) auxiliary-module fitting for $R^{(L\rightarrow k)}$ and $W^{(g)}$, (ii) empirical-CDF fitting, and (iii) final threshold calibration. The score function, including the auxiliary modules and CDFs, is fixed before the threshold-calibration set is scored. Given $n$ exchangeable benign calibration scores $S_1,\ldots,S_n$ and desired FPR $\alpha$, let
\begin{equation}
k=\left\lceil(n+1)(1-\alpha)\right\rceil,\qquad \tau_\alpha=S_{(k)},
\end{equation}
where $S_{(k)}$ is the $k$-th order statistic (and $\tau_\alpha=+\infty$ if $k>n$). We flag a future input iff $S_{n+1}>\tau_\alpha$. Under exchangeability between calibration and future benign scores, this rule controls the marginal FPR at $\alpha$; the proof and finite-sample study are in \autoref{app:calibration_failure}. The guarantee is marginal over calibration sets, has resolution $1/(n+1)$, and does not hold under distribution shift. For a changed classifier, preprocessing pipeline, or deployment population, target-domain benign data are required for recalibration; if the CDFs are refitted, a separate target-domain threshold split preserves the independence condition.
\label{sec3}

\section{Experiments}\label{sec4}
\vspace{-1mm}

\begin{table*}[t]
\centering
\caption{The AUC of different adversarial detection approaches on CIFAR-10. The results are the mean and standard deviation of 5 runs. Our methods are included for comparison. Classifier: ResNet110, FGSM: $\epsilon = 0.05$, PGD: $\epsilon = 0.02$. Note that our methods and BEYOND need no AE for training, leading to the same value on both seen and unseen settings. The bolded values are the best performance, and the underlined italicized values are the second-best performance.}
\label{tab:auc_ours_vs_baselines}
\resizebox{\textwidth}{!}{%
\begin{tabular}{lcccccccccc}
\toprule
\textbf{Method} & \multicolumn{4}{c}{\textbf{Unseen (Attacks in training are excluded from tests)}} & \multicolumn{5}{c}{\textbf{Seen (Attacks in training are included in tests)}} \\
\cmidrule(lr){2-5} \cmidrule(lr){6-10}
 & \textbf{FGSM} & \textbf{PGD} & \textbf{AutoAttack} & \textbf{Square} & \textbf{FGSM} & \textbf{PGD} & \textbf{CW} & \textbf{AutoAttack} & \textbf{Square} \\
\midrule
DkNN~\citep{papernot2018deep} & 61.55 $_{\pm0.023}$ & 51.22 $_{\pm0.026}$ & 52.12 $_{\pm0.023}$ & 59.46 $_{\pm0.022}$ 
     & 61.55 $_{\pm0.023}$ & 51.22 $_{\pm0.026}$ & 61.52 $_{\pm0.028}$ & 52.12 $_{\pm0.023}$ & 59.46 $_{\pm0.022}$ \\
kNN~\citep{dubey2019defense}  & 61.83 $_{\pm0.018}$ & 54.52 $_{\pm0.022}$ & 52.67 $_{\pm0.022}$ & 73.39 $_{\pm0.020}$ 
     & 61.83 $_{\pm0.018}$ & 54.52 $_{\pm0.022}$ & 62.23 $_{\pm0.019}$ & 52.67 $_{\pm0.022}$ & 73.39 $_{\pm0.020}$ \\
LID~\citep{ma2018characterizing}  & 71.08 $_{\pm0.024}$ & 61.33 $_{\pm0.025}$ & 55.56 $_{\pm0.021}$ & 66.18 $_{\pm0.025}$ 
     & 73.61 $_{\pm0.020}$ & 67.98 $_{\pm0.020}$ & 55.68 $_{\pm0.021}$ & 56.33 $_{\pm0.024}$ & 85.94 $_{\pm0.018}$ \\
\citet{hu2019new}   & 84.51 $_{\pm0.025}$ & 58.59 $_{\pm0.028}$ & 53.55 $_{\pm0.029}$ & 95.82 $_{\pm0.020}$ 
     & 84.51 $_{\pm0.025}$ & 58.59 $_{\pm0.028}$ & 91.02 $_{\pm0.022}$ & 53.55 $_{\pm0.029}$ & 95.82 $_{\pm0.020}$ \\
\citet{mao2021adversarial}  & 95.33 $_{\pm0.012}$ & 82.61 $_{\pm0.016}$ & 81.95 $_{\pm0.020}$ & 85.76 $_{\pm0.019}$ 
     & 95.33 $_{\pm0.012}$ & 82.61 $_{\pm0.016}$ & 83.10 $_{\pm0.018}$ & 81.95 $_{\pm0.020}$ & 85.76 $_{\pm0.019}$ \\
LNG~\citep{abusnaina2021adversarial}  & 98.51 & 63.14 & 58.47 & 94.71 
     & \underline{\textit{99.88}} & 91.39 & 89.74 & 84.03 & \underline{\textit{98.82}} \\
BEYOND~\citep{he2024beyond} & 98.89 $_{\pm0.013}$ & \underline{\textit{99.28}} $_{\pm0.020}$ & 99.16 $_{\pm0.021}$ & \textbf{99.27} $_{\pm0.016}$ 
       & 98.89 $_{\pm0.013}$ & \underline{\textit{99.28}} $_{\pm0.020}$ & 99.20 $_{\pm0.008}$ & 99.16 $_{\pm0.021}$ & \textbf{99.27} $_{\pm0.016}$ \\
\midrule
\textbf{Our Approaches} & & & & & & & & & \\
RT  & \textbf{99.93} $_{\pm0.005}$ & 96.89 $_{\pm0.071}$ & \textbf{99.99} $_{\pm0.000}$ & 85.38 $_{\pm0.344}$ 
    & \textbf{99.93} $_{\pm0.005}$ & 96.89 $_{\pm0.071}$ & \textbf{99.99} $_{\pm0.002}$ & \textbf{99.99} $_{\pm0.000}$ & 85.38 $_{\pm0.344}$ \\
LT  & 97.50 $_{\pm0.038}$ & 98.61 $_{\pm0.042}$ & \underline{\textit{99.60}} $_{\pm0.018}$ & \underline{\textit{97.47}} $_{\pm0.036}$ 
    & 97.50 $_{\pm0.038}$ & 98.61 $_{\pm0.042}$ & 97.08 $_{\pm0.027}$ & \underline{\textit{99.60}} $_{\pm0.018}$ & 97.47 $_{\pm0.036}$ \\
RLT  & \underline{\textit{99.85}} $_{\pm0.005}$ & \textbf{99.37} $_{\pm0.011}$ & \textbf{99.99} $_{\pm0.000}$ & 95.95 $_{\pm0.102}$ 
    & 99.85 $_{\pm0.005}$ & \textbf{99.37} $_{\pm0.011}$ & \underline{\textit{99.91}} $_{\pm0.004}$ & \textbf{99.99} $_{\pm0.000}$ & 95.95 $_{\pm0.102}$ \\
\bottomrule
\end{tabular}
}
\end{table*}

\begin{table*}[t]
\centering
\caption{The AUC of different adversarial detection approaches on ImageNet. The results are the mean and standard deviation of 5 runs. Our methods are included for comparison. To align with baselines, classifier: DenseNet121, FGSM: $\epsilon=0.05$, PGD: $\epsilon=0.02$ are used. Due to memory and resource constraints, baseline methods are not evaluated against AutoAttack on ImageNet.}
\label{tab:auc_ours_vs_baselines_imagenet}
\resizebox{0.75\textwidth}{!}{%
\begin{tabular}{lccccc}
\toprule
\textbf{Method} & \multicolumn{2}{c}{\textbf{Unseen}} & \multicolumn{3}{c}{\textbf{Seen}} \\
\cmidrule(lr){2-3} \cmidrule(lr){4-6}
 & \textbf{FGSM} & \textbf{PGD} & \textbf{FGSM} & \textbf{PGD} & \textbf{CW} \\
\midrule
DkNN~\citep{papernot2018deep} & 89.16 $_{\pm0.038}$ & 78.00 $_{\pm0.041}$ & 89.16 $_{\pm0.038}$ & 78.00 $_{\pm0.041}$ 
     & 68.91 $_{\pm0.044}$ \\
kNN~\citep{dubey2019defense}  & 51.63 $_{\pm0.04}$ & 51.14 $_{\pm0.039}$ & 51.63 $_{\pm0.04}$ & 51.14 $_{\pm0.039}$ 
     & 50.73 $_{\pm0.04}$ \\
LID~\citep{ma2018characterizing}  & 90.32 $_{\pm0.046}$ & 52.56 $_{\pm0.038}$ & \underline{\textit{99.24}} $_{\pm0.043}$ & 98.09 $_{\pm0.042}$ 
     & 58.83 $_{\pm0.041}$ \\
\citet{hu2019new}   & 72.56 $_{\pm0.037}$ & 86.00 $_{\pm0.042}$ & 72.56 $_{\pm0.037}$ & 86.00 $_{\pm0.042}$ 
     & 80.79 $_{\pm0.044}$ \\
LNG~\citep{abusnaina2021adversarial}  & 96.85 & 89.61 & \textbf{99.53} & \underline{\textit{98.42}} 
     & 86.05 \\
BEYOND~\citep{he2024beyond} & \underline{\textit{97.59}} $_{\pm0.04}$ & 96.26 $_{\pm0.045}$ & 97.59 $_{\pm0.04}$ & 96.26 $_{\pm0.045}$ 
       & \textbf{95.46} $_{\pm0.047}$ \\
\midrule
\textbf{Our Approaches} & & & & & \\
RT  & 94.31 $_{\pm0.457}$ & \textbf{99.99} $_{\pm0.000}$ & 94.31 $_{\pm0.457}$ & \textbf{99.99} $_{\pm0.000}$
    & 92.18 $_{\pm0.135}$  \\
LT  & 96.18 $_{\pm0.028}$ & \underline{\textit{97.89}} $_{\pm0.021}$ & 96.18 $_{\pm0.028}$ & 97.89 $_{\pm0.021}$
    & \underline{\textit{94.06}} $_{\pm0.215}$ \\
RLT  & \textbf{97.60} $_{\pm0.048}$ & \textbf{99.99} $_{\pm0.000}$ & 97.60 $_{\pm0.048}$ & \textbf{99.99} $_{\pm0.000}$
    & 91.19 $_{\pm0.022}$ \\
\bottomrule
\end{tabular}
}
\vspace{-4mm}
\end{table*}

\begin{table}[h]
\centering
\caption{System RA (\%) under Orthogonal-PGD on CIFAR-10/ResNet-110.}
\label{tab:adaptive_attack}
\resizebox{0.7\textwidth}{!}{
\begin{tabular}{lcccc}
\toprule
\textbf{Defense} & \multicolumn{2}{c}{$\boldsymbol{L_\infty=0.01}$} & \multicolumn{2}{c}{$\boldsymbol{L_\infty=8/255}$} \\
\cmidrule(lr){2-3} \cmidrule(lr){4-5}
 & \textbf{RA@FPR5\%} & \textbf{RA@FPR50\%} & \textbf{RA@FPR5\%} & \textbf{RA@FPR50\%} \\
\midrule
\textbf{Ours (RLT)}     & \underline{\textit{75.40}} & \textbf{99.58} & \textbf{33.70} & \textbf{80.77} \\
BEYOND~\citep{he2024beyond}         & \textbf{88.38} & \underline{\textit{98.81}} & \underline{\textit{13.80}} & \underline{\textit{48.20}} \\
Trapdoor~\cite{shan2020gotta}                & 0.00  & 7.00  & 0.00  & 8.00  \\
DLA~\cite{sperl2020dla}                     & 62.60 & 83.70 & 0.00  & 28.20 \\
SID~\cite{tian2021detecting}                     & 6.90  & 23.40 & 0.00  & 1.60  \\
SPAM~\cite{liu2019detection}                    & 1.20  & 46.00 & 0.00  & 38.00 \\
\bottomrule
\end{tabular}
}
\end{table}

\begin{table}[t]
\centering
\caption{System RA (\%) under end-to-end PGD + BPDA at $\epsilon=8/255$ (ResNet-110, CIFAR-10). Among source inputs correctly classified before attack, RA is one minus the joint misclassification-and-evasion rate. Results are mean and standard deviation over five runs.}
\vspace{0.5mm}
\label{tab:bpda_pgd}
\resizebox{1.0\textwidth}{!}{
\begin{tabular}{lccccccccccc}
\toprule
$\boldsymbol{\lambda}$ & \textbf{@FPR5\%} & \textbf{@FPR10\%} & \textbf{@FPR15\%} & \textbf{@FPR20\%} & \textbf{@FPR25\%} & \textbf{@FPR30\%} & \textbf{@FPR35\%} & \textbf{@FPR40\%} & \textbf{@FPR45\%} & \textbf{@FPR50\%} \\
\midrule
1.00 & $20.72_{\pm8.04}$ & $25.96_{\pm6.88}$ & $32.05_{\pm5.72}$ & $38.23_{\pm5.11}$ & $44.48_{\pm4.52}$ & $50.15_{\pm4.13}$ & $55.15_{\pm3.56}$ & $60.09_{\pm3.10}$ & $64.54_{\pm2.42}$ & $68.53_{\pm2.21}$ \\
0.50 & $22.64_{\pm4.46}$ & $30.45_{\pm3.25}$ & $37.96_{\pm2.08}$ & $44.78_{\pm1.81}$ & $51.19_{\pm1.30}$ & $56.41_{\pm0.98}$ & $61.21_{\pm0.96}$ & $65.44_{\pm0.96}$ & $69.35_{\pm0.83}$ & $73.00_{\pm0.92}$ \\
0.25 & $23.39_{\pm1.76}$ & $32.92_{\pm0.49}$ & $40.88_{\pm1.04}$ & $48.04_{\pm1.13}$ & $54.32_{\pm1.58}$ & $59.37_{\pm1.44}$ & $63.90_{\pm1.82}$ & $67.98_{\pm1.82}$ & $71.51_{\pm1.89}$ & $74.83_{\pm1.85}$ \\
\midrule
\citet{mao2021adversarial} & & & & & 18.97 & & & & & \\
BEYOND~\citep{he2024beyond} & & & & & 19.45 & & & & & \\
\bottomrule
\end{tabular}
}
\end{table}

\begin{table}[t]
\vspace{-4mm}
\centering
\caption{System RA (\%) under end-to-end PGD + BPDA at $\epsilon=8/255$ with an ATC (ResNet-110, CIFAR-10). Results are mean and standard deviation over five runs.}
\vspace{0mm}
\label{tab:bpda_pgd_wATC}
\resizebox{1.0\textwidth}{!}{%
\begin{tabular}{lccccccccccc}
\toprule
$\boldsymbol{\lambda}$ & \textbf{@FPR5\%} & \textbf{@FPR10\%} & \textbf{@FPR15\%} & \textbf{@FPR20\%} & \textbf{@FPR25\%} & \textbf{@FPR30\%} & \textbf{@FPR35\%} & \textbf{@FPR40\%} & \textbf{@FPR45\%} & \textbf{@FPR50\%} \\
\midrule
1.00 & $93.32_{\pm0.13}$ & $93.38_{\pm0.14}$ & $93.41_{\pm0.11}$ & $93.43_{\pm0.09}$ & $93.97_{\pm0.12}$ & $94.49_{\pm0.15}$ & $94.99_{\pm0.14}$ & $95.50_{\pm0.13}$ & $95.93_{\pm0.14}$ & $96.32_{\pm0.11}$ \\
0.50 & $93.53_{\pm0.15}$ & $94.17_{\pm0.12}$ & $94.27_{\pm0.11}$ & $94.33_{\pm0.09}$ & $94.86_{\pm0.13}$ & $95.41_{\pm0.09}$ & $95.88_{\pm0.09}$ & $96.27_{\pm0.11}$ & $96.65_{\pm0.14}$ & $97.00_{\pm0.15}$ \\
0.25 & $93.57_{\pm0.14}$ & $94.46_{\pm0.16}$ & $94.60_{\pm0.13}$ & $94.85_{\pm0.07}$ & $95.40_{\pm0.05}$ & $95.87_{\pm0.08}$ & $96.32_{\pm0.14}$ & $96.71_{\pm0.12}$ & $97.04_{\pm0.12}$ & $97.30_{\pm0.14}$ \\
\midrule
\citet{mao2021adversarial} & & & & & 75.09 & & & & & \\
BEYOND~\citep{he2024beyond} & & & & & 93.20 & & & & & \\
\bottomrule
\end{tabular}
}
\end{table}

\begin{figure}[h]
\vspace{-1mm}
    \centering
    \includegraphics[width=1.0\linewidth]{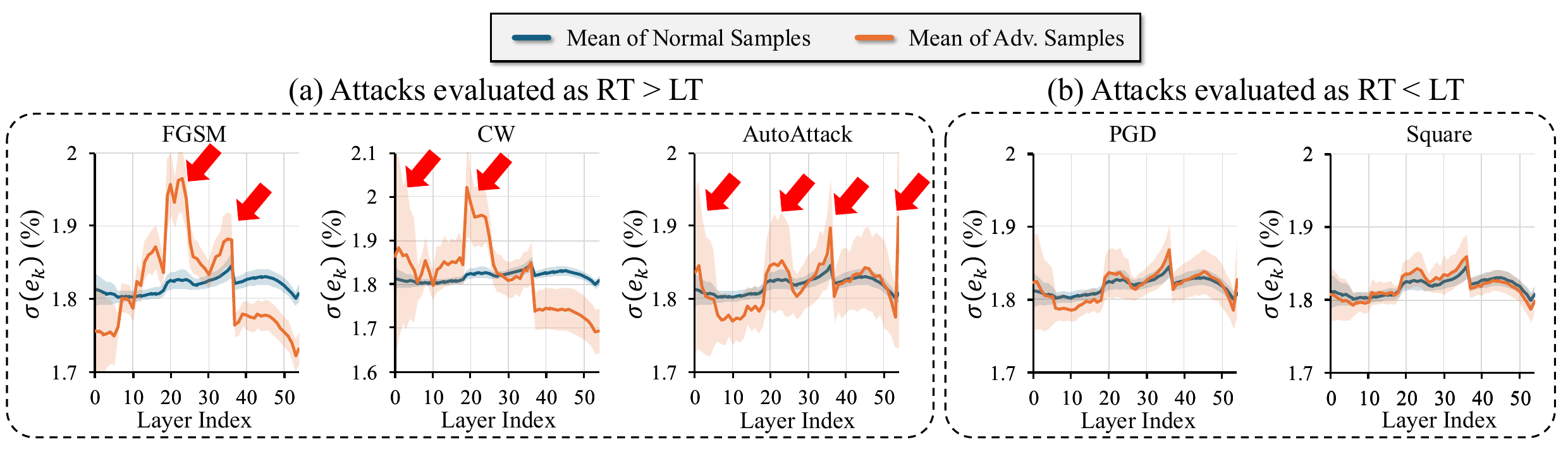}
    \vspace{-8mm}
    \caption{
       \textbf{Distribution of RT reconstruction errors} across the recovered layers on CIFAR-10/ResNet-110. The errors are normalized across layers as $\sigma(e_{RT}(x))$.
    }
    \label{fig:error_distribution}
\vspace{-5mm}
\end{figure}

\subsection{Experimental Setup}
\label{exp_4_1}

\textbf{Datasets.} We evaluate our method on CIFAR-10, CIFAR-100, and ImageNet. CIFAR-10 and CIFAR-100 each contain $60{,}000$ $32\times32$ images with standard training (50k) and test (10k) splits, across 10 and 100 classes respectively. For ImageNet, we use the official training and validation sets, resizing all images to $256\times256$ and applying standard normalization. CIFAR-10 is used for the main comparison to prior detectors, CIFAR-100 evaluates fine-grained class scaling, and ImageNet serves as the large-scale benchmark.

\textbf{Backbone Models.} For the main detector comparison, we follow the baseline protocols and use ResNet-110 for CIFAR-10 and DenseNet-121 for ImageNet. The expanded AFLS analysis uses ResNet-18, ResNet-34, VGG-11, and VGG-13 across CIFAR-10, CIFAR-100, and ImageNet. We further evaluate architectural transfer with MobileNet-V2, ShuffleNet-V2, RepVGG-a0, and a supervised ViT-B/16 (\autoref{sec:arch_eval} and \autoref{app:vit}).

\textbf{Threat Models.} We evaluate our detection framework under two standard adversarial settings: \emph{Limited Knowledge} and \emph{Perfect Knowledge}, following the protocol of previous works~\cite{apruzzese2023real, he2024beyond}. In the \emph{Limited Knowledge} setting, the adversary has full access to the target classifier but is unaware of the detection mechanism, which remains confidential. In contrast, the \emph{Perfect Knowledge} (adaptive attack) setting assumes that the adversary has full knowledge of both the classifier and the detection strategy, enabling it to craft attacks specifically to evade detection.

\textbf{Attack Methods.} We evaluate robustness and detection performance under a diverse suite of standard attacks, including Fast Gradient Sign Method (FGSM), Projected Gradient Descent (PGD), Carlini--Wagner (CW), AutoAttack, and Square Attack~\cite{andriushchenko2020square}. To assess adaptive evasion, we also include Orthogonal-PGD~\cite{bryniarski2021evading} and end-to-end PGD objectives that explicitly combine misclassification with detector-score suppression. We report each adaptive objective, perturbation budget, and operating point in \autoref{exp2}, with complete configurations in \autoref{app:attack_configurations}.

\textbf{Implementation Details.} All models are implemented in PyTorch. The main experiments use batch size 32 and AdamW with learning rate $10^{-4}$ and weight decay $0.01$; recovery modules and transformation matrices are fitted for 50 epochs on benign data. Auxiliary-module fitting, empirical-CDF fitting, and final threshold calibration use disjoint benign roles as specified in \autoref{sec:calibration_protocol}. The target classifier remains~fixed.

\vspace{-2mm}
\subsection{Detection Performance under Standard Attacks}
\vspace{-2mm}
\label{exp1}

We evaluate the effectiveness of our proposed detection scores, namely RT, LT, and RLT, on CIFAR-10 and ImageNet under standard adversarial threat models, including FGSM, PGD, CW, AutoAttack, and Square Attack. We benchmark against established baselines such as LID~\cite{ma2018characterizing}, DkNN~\cite{papernot2018deep}, LNG~\cite{abusnaina2021adversarial}, and the recent SSL-based BEYOND~\cite{he2024beyond}. The Area Under the Receiver Operating Characteristic Curve (ROC-AUC) results, which measure detection performance over thresholds, are reported in \autoref{tab:auc_ours_vs_baselines} for CIFAR-10 and \autoref{tab:auc_ours_vs_baselines_imagenet} for ImageNet. Additional CIFAR-100 and architecture-generalization results are provided in \autoref{sec:arch_eval} and \autoref{app:additional_afls_validation}.

On CIFAR-10, RT has higher AUC than LT for FGSM, CW, and AutoAttack, while LT has higher AUC for PGD and Square Attack. RLT achieves the best or second-best result for FGSM, PGD, CW, and AutoAttack; for Square Attack, LT performs better than RLT. These differences show why combining RT and LT is useful across the reported attacks, without assigning a particular attack to a specific layer. RLT uses only signals from the target classifier and requires neither adversarial examples nor an external pretrained model for fitting.

On ImageNet, RLT has the highest unseen-FGSM AUC (97.60), while it is below the strongest baselines for seen FGSM. For PGD, RT and RLT both reach 99.99 AUC in the seen and unseen settings, and LT is second-best for seen CW with 94.06 AUC (\autoref{tab:auc_ours_vs_baselines_imagenet}). These results show that signals fitted from benign representations of the target model remain informative in this experiment, while the auxiliary modules and thresholds must still be fitted for each target model.

\subsection{Detection Performance under Adaptive Attacks}
\label{exp2}

To reduce the risk of gradient-obfuscation artifacts and provide end-to-end gradient flow through our detection framework, we apply the Backward Pass Differentiable Approximation (BPDA)~\cite{athalye2018obfuscated} to components that may otherwise block gradients, such as the empirical-CDF normalization.

Among inputs that the classifier predicts correctly before attack, we define classification attack success as misclassification, conditional evasion as a detector score at or below threshold among successful attacks, and joint success as both events. System RA is one minus joint success; detailed definitions are in \autoref{app:adaptive_scope}. We first evaluate Orthogonal-PGD following BEYOND~\cite{he2024beyond}, with full knowledge of the classifier and detector.
\autoref{tab:adaptive_attack} reports RA at two $L_\infty$ budgets ($0.01$ and $8/255$) and operating points (\mbox{FPR=5\%} and \mbox{FPR=50\%}). Relative to BEYOND, RLT is higher at three of the four budget/operating-point combinations. At $\epsilon=0.01$ and FPR 5\%, BEYOND is higher (88.38\% versus 75.40\%). We next evaluate a fully end-to-end objective that directly suppresses the fused RLT score.

We further evaluate fully end-to-end PGD attacks on the fused RLT score using the untargeted minimization objective
$\min_{\|\delta\|_\infty\le \epsilon}\{-\mathcal{L}_{\mathrm{cls}}(x+\delta,y)+\lambda\cdot\mathrm{RLT}(x+\delta)\}$.
The first term induces misclassification and the second suppresses the detector. For $\lambda\in\{0.25,0.50,1.00\}$, standalone RA at FPR 5\% is $20.72$--$23.39\%$ (\autoref{tab:bpda_pgd}), showing that the detector alone is substantially stressed by this white-box attack. At FPR 25\%, its $44.48$--$54.32\%$ RA exceeds the reported Mao and BEYOND values of $18.97\%$ and $19.45\%$. When combined with an ATC, RA is $93.32$--$93.57\%$ at FPR 5\% (\autoref{tab:bpda_pgd_wATC}). The $\lambda$ sweep quantifies sensitivity to the balance between misclassification and detector suppression.

We also measure the input gradients of the adaptive objectives on $1{,}000$ successful PGD examples (\autoref{tab:gradient_alignment}). The misclassification and RT-suppression gradients have median cosine $-0.762$ and negative cosine on 99.8\% of samples, indicating a strong local trade-off. The RT--LT median is $-0.053$ with an interquartile range spanning zero, showing that their relative gradient direction varies across samples.

\noindent
\begin{minipage}[t]{0.50\textwidth}
\vspace{-2mm}
As a gradient-masking check, we evaluate the gradient-free SimBA attack~\cite{guo2019simple} with 1,000 classifier-output queries. It lowers undefended RA to 5.54\%, while the RLT system has 97.62\% RA at FPR 5\% (\autoref{tab:simba}). This classifier-output attack complements the white-box study by testing the prediction path without first-order gradients.
\end{minipage}\hfill
\begin{minipage}[t]{0.47\textwidth}
\vspace{-5mm}
\centering
\captionof{table}{System RA (\%) under a 1,000-query classifier-output SimBA attack at $\epsilon=8/255$ (ResNet-110, CIFAR-10).}
\vspace{1mm}
\label{tab:simba}
\resizebox{0.8\linewidth}{!}{%
\begin{tabular}{c|cc}
\toprule
\textbf{Defense} & \textbf{RA@FPR5\%} & \textbf{RA@FPR50\%} \\
\midrule
None & \multicolumn{2}{c}{\textbf{5.54}}  \\
\textbf{Ours (RLT)} & \textbf{97.62} & \textbf{97.85} \\
\bottomrule
\end{tabular}
}
\end{minipage}
\par

\subsection{Empirical Evaluation of the Proposed Assumption}
\label{exp3}

We first examine how the RT reconstruction errors are distributed across recovered layers on CIFAR-10/ResNet-110 under FGSM, PGD, CW, AutoAttack, and Square Attack. In \autoref{fig:error_distribution}, benign inputs have relatively even error magnitudes across layers, whereas several attacks produce one or a few pronounced peaks. This pattern motivates the layer-concentration term in RT.

We directly evaluate AFLS in \autoref{app:additional_afls_validation} by comparing each clean input with a benignly transformed or attack-perturbed version. Across 12 dataset/model combinations, the mean probability that at least one $A_k$ exceeds its benign reference bound is 3.57\% for random crop, horizontal flip, and random noise, compared with 87.51\% for FGSM, PGD, CW, and AutoAttack. For attack-generated inputs, the largest $A_k$ accounts for 70.17\% of the total on average. The accompanying LT comparison further identifies CW and AutoAttack inputs for which every $A_k$ remains within its benign bound.

We also analyze calibration of the learned RLT detector in \autoref{app:calibration_failure}. A disjoint 5,000/5,000 threshold-calibration/evaluation split gives 5.58\% held-out FPR at nominal 5\% (95\% Wilson interval $[4.98,6.25]\%$), and repeated subsampling shows decreasing threshold variability as calibration size grows. Under severe benign distribution shifts, a threshold calibrated on the source distribution can over-flag inputs; recalibration on representative target-domain benign samples restores FPR to $4.50$--$5.36\%$ across five tested shifts.

\vspace{-3mm}
\subsection{Implementation Costs}
\label{exp4}
\vspace{-3mm}

\noindent
\begin{minipage}[t]{0.56\textwidth}
\vspace{-4mm}
\centering
\captionof{table}{Comparison of implementation costs in terms of FLOPs, parameters, and model size overhead when the target model is ResNet110 with the CIFAR-10 dataset.}
\label{tab:implementation_cost}
\vspace{1mm}
\resizebox{\linewidth}{!}{%
\begin{tabular}{lccc}
\toprule
\textbf{Method} & \textbf{FLOPs (G)} & \textbf{Params (M)} & \textbf{Model Size Overhead ($\boldsymbol{\times}$)} \\
\midrule
\citet{mao2021adversarial} & 5.25 & 38.12 & 22.02 \\
LNG~\citep{abusnaina2021adversarial} & \textbf{0.286} & \underline{\textit{8.33}} & \underline{\textit{4.81}} \\
BEYOND~\citep{he2024beyond} & 0.715 & 20.62 & 11.91 \\
\textbf{Ours (RLT)} & \underline{\textit{0.491}} & \textbf{2.59} & \textbf{1.49} \\
\bottomrule
\end{tabular}
}
\end{minipage}\hfill
\begin{minipage}[t]{0.41\textwidth}
\vspace{0pt}
We evaluate RLT's implementation cost under the same configuration as in \autoref{tab:auc_ours_vs_baselines}. As shown in \autoref{tab:implementation_cost}, RLT has lower FLOPs, parameter count, and model-size overhead than Mao et al. and BEYOND. Compared with LNG, RLT uses fewer parameters (2.59M versus 8.33M) and less model-size overhead
\end{minipage}
 (1.49$\times$ versus 4.81$\times$), but more FLOPs (0.491G versus 0.286G). A detailed analysis across target architectures is provided in \autoref{app:overhead}.
\par

\section{Conclusion}
We introduced an adversarial example detector based on the target model's internal representations and motivated by the A Few Large Shifts empirical hypothesis. AFLS connects localized growth in representation changes to layer-wise Lipschitz continuity. RT measures intermediate-layer reconstruction inconsistency, LT measures output changes under learned transformations relative to intermediate-feature changes, and RLT combines their scores after normalization on benign data. Across CIFAR-10, CIFAR-100, and ImageNet, RLT achieves strong detection under standard attacks with low implementation overhead, while the adaptive, architectural-transfer, calibration, and system-level studies characterize its behavior across complementary evaluation settings.

\textbf{Limitations and Future Work.} AFLS is an empirical, setting-dependent hypothesis rather than a universal property, and RLT is a detector rather than a certified defense. Standalone RLT is substantially stressed by the end-to-end BPDA attack at low FPR. The adaptive evaluation covers Orthogonal-PGD, a finite $\lambda$ sweep for end-to-end PGD, and a classifier-output SimBA attack; stronger detector-aware optimization, restart strategies, and query attacks remain important directions. The gradient diagnostics identify a strong local classification--RT trade-off, while RT--LT alignment varies across samples.

The finite-sample threshold rule controls marginal benign FPR only when calibration and future benign scores are exchangeable. Source thresholds can over-flag shifted benign inputs, so representative target-domain recalibration is required after changes to the classifier, data, or preprocessing. The architectural evidence covers several supervised CNN families and one supervised ViT-B/16 configuration; DINO-, MAE-, and other substantially different training paradigms are outside the validated scope.
\label{sec6}

\section*{Acknowledgments}

This work was supported in part by the DARPA Young Faculty Award, the National Science Foundation (NSF) under Grants \#2127780, \#2319198, \#2321840, \#2312517, and \#2235472, the Semiconductor Research Corporation (SRC), the Office of Naval Research through the Young Investigator Program Award, and Grants \#N00014-21-1-2225 and N00014-24-1-2547, Army Research Office Grant \#W911NF2410360. Additionally, support was provided by the Air Force Office of Scientific Research under Award \#FA9550-22-1-0253.

\bibliography{main}
\bibliographystyle{tmlr}
\newpage
\appendix
\section*{Appendix}
\renewcommand{\sectionautorefname}{Appendix}
\renewcommand{\subsectionautorefname}{\sectionautorefname}

\section{Related Work and Background}\label{sec2}
\vspace{-2mm}
\paragraph{Input Transformation-Based Detectors.}
A well-established line of work detects adversarial examples by applying simple input transformations and monitoring the model's response. For instance, \citet{xu2017feature} proposed \emph{feature squeezing}, which reduces input precision or applies spatial smoothing to suppress adversarial noise. \citet{liang2018detecting} used quantization and denoising techniques, while other works explore random crops or region replacements. These preprocessing-based detectors are lightweight and model-agnostic but assume that perturbations are fragile to such transformations. Their effectiveness often hinges on tuning a detection threshold, and their robustness degrades against adaptive attacks that are explicitly trained to remain invariant under the applied transformations. Coarse transformations may also degrade clean-data performance, increasing false positives.
\vspace{-3mm}
\paragraph{Feature Statistics-Based Detectors.}
Another category of detectors analyzes statistical anomalies in the network's hidden representations. \citet{ma2018characterizing} proposed Local Intrinsic Dimensionality (LID) to measure how local feature neighborhoods expand under adversarial perturbations. \citet{feinman2017detecting} leveraged kernel density estimation and uncertainty measures, while \citet{lee2018simple} introduced a Mahalanobis distance-based detector by modeling class-conditional distributions in intermediate layers. These approaches rely on hand-crafted features from latent activations and often require storing high-dimensional embeddings or computing pairwise distances, which can be computationally intensive. Moreover, many methods are vulnerable to adaptive attacks that mimic the distributional properties of benign inputs.
\vspace{-3mm}
\paragraph{Model Behavior-Based Detectors.}
Some methods probe the internal behavior of the model itself to identify inconsistencies caused by adversarial perturbations. \citet{lu2017safetynet} encoded final-layer activations into binary vectors for SVM-based detection, while \citet{metzen2017detecting} appended auxiliary classifiers to intermediate layers. Others, such as \citet{ma2019nic} and \citet{carrara2018adversarial}, focused on monitoring neuron activation paths or ensemble agreement. These methods often require additional training, architectural modifications, or model ensembles, increasing overhead and complicating deployment. Many also depend on adversarial examples for supervision, which limits their generalizability to unseen attacks.
\vspace{-3mm}
\paragraph{Self-Supervised and Consistency-Based Detectors.}
Recent approaches have explored feature consistency using self-supervised learning (SSL) models. For example, BEYOND~\citep{he2024beyond} detects adversarial inputs by measuring feature stability across augmentations in a large SSL encoder. These methods eliminate the need for adversarial examples and external graph structures but introduce high computational costs due to reliance on large pretrained models; in our ResNet-110 comparison, BEYOND adds 11.91$\times$ the target classifier's parameter count. Additionally, they often require access to domain-specific SSL models and retraining or fine-tuning of auxiliary heads.
\vspace{-3mm}
\paragraph{Our Approach.}
In contrast to the methods above, our framework evaluates the consistency of feature transformations across successive layers within the target network. After benign-only fitting and calibration, the detector analyzes these model-internal transformations for a single test input, without a query-time reference set, external encoder, or extensive augmentation. This approach is grounded in our \textbf{A Few Large Shifts Assumption}, which connects adversarial perturbations with large, localized representation changes across a small number of consecutive layers.

\section{Theoretical Analysis}
\label{app:assA}

We provide sufficient-condition analyses for RT, LT, and RLT. The results characterize how each score responds under explicit conditions on intermediate representations, learned transformations, and normalized score margins. We evaluate the empirical AFLS hypothesis separately in Section~\ref{exp3} and \autoref{app:additional_afls_validation}.

\subsection{Layer-wise Lipschitz Interpretation}
\label{app:lipschitz}

Our layer-wise Lipschitz framing applies local Lipschitz reasoning separately to each transformation in the network. Let $f_k$ denote the $k$-th layer, so that $\mathbf z_k=f_k(\mathbf z_{k-1})$. A Lipschitz constant $L_k(\mathcal N)$ on a neighborhood $\mathcal N$ is any value satisfying
\begin{equation}
    \|f_k(\mathbf{z}'_{k-1}) - f_k(\mathbf{z}_{k-1})\| \le L_k \|\mathbf{z}'_{k-1} - \mathbf{z}_{k-1}\|
\end{equation}
for all $\mathbf z_{k-1},\mathbf z'_{k-1}\in\mathcal N$. For any particular pair in $\mathcal N$, the observed ratio is bounded above by $L_k(\mathcal N)$:
\begin{equation}
\frac{\|f_k(\mathbf z'_{k-1})-f_k(\mathbf z_{k-1})\|_2}
{\|\mathbf z'_{k-1}-\mathbf z_{k-1}\|_2}\leq L_k(\mathcal N).
\end{equation}
For the AFLS quantity in Section~\ref{sec:afls_hypothesis}, whenever $\|z_k(x')-z_k(x)\|_2>0$,
\begin{equation}
\sqrt{A_k(x,x')\frac{D_{k+1}}{D_k}}
\leq
\frac{\|z_{k+1}(x')-z_{k+1}(x)\|_2}{\|z_k(x')-z_k(x)\|_2}
\leq L_{k+1}(\mathcal N).
\end{equation}
The first inequality follows because $\varepsilon_A$ enlarges the denominator of $A_k$. Thus, a large dimension-corrected $A_k$ gives a lower bound on any Lipschitz constant valid for the neighborhood containing the observed pair. The middle quantity describes the change along the observed pair, whereas $L_{k+1}(\mathcal N)$ uniformly bounds all pairs in the neighborhood. Comparing $A_k$ with its benign reference $b_k$ therefore identifies transitions whose observed amplification is unusually large relative to benign layer-wise behavior.

At inference time, RT evaluates the same layer-wise intuition from a single query by using recovery modules trained on benign data. The next result gives a sufficient condition under which its reconstruction error increases.

\subsection{Justification for Layer-wise Reconstruction Error in RT}

We first derive a sufficient condition under which an intermediate recovery residual increases relative to a benign reference.

\begin{assumption}[Approximate Invertibility]\label{assump:inv}
For each intermediate layer $i \in \{k_{RT}, \dots, L-1\}$, the fitted recovery function $R^{(L \rightarrow i)}$ satisfies, for benign inputs $x$,

\begin{equation}
\| z_i(x) - R^{(L \rightarrow i)}(z_L(x)) \| \le \varepsilon,
\end{equation}
for some small constant $\varepsilon > 0$.
\end{assumption}

\begin{assumption}[Lipschitz Continuity of the Recovery Function]\label{assump:flat}
For the relevant final-layer representations, the fitted recovery function $R^{(L \rightarrow i)}$ is Lipschitz continuous with constant $\alpha \ll 1$:

\begin{equation}
\| R^{(L \rightarrow i)}(z'_L) - R^{(L \rightarrow i)}(z_L) \| \le \alpha \cdot \| z'_L - z_L \|.
\end{equation}
\end{assumption}

\begin{theorem}[Sufficient Condition for an Increased RT Residual]\label{thm:rec}
Let $r_i(u)=\|z_i(u)-R^{(L\rightarrow i)}(z_L(u))\|_2$. Suppose \autoref{assump:inv} holds for benign $x$, and let $x'=x+\delta$. If
\begin{equation}
\|z_i(x')-z_i(x)\|_2>
2\varepsilon+\|R^{(L\rightarrow i)}(z_L(x'))-R^{(L\rightarrow i)}(z_L(x))\|_2,
\end{equation}
then
\begin{equation}
r_i(x')^2>r_i(x)^2.
\end{equation}
\end{theorem}

(Proof: \autoref{pro:theorem1})

\autoref{assump:flat} controls the recovery-map term by $\alpha\|z_L(x')-z_L(x)\|_2$. Together with the strict margin in \autoref{thm:rec}, this yields an increased recovery residual.

\subsection{Justification for Ratio-based Logit-layer Deviation in LT}

LT measures the output change under a learned transformation relative to the corresponding change in intermediate features. The following results characterize the output-instability condition captured by this ratio.

\begin{assumption}[Benign Augmentation Stability]\label{assump:small_aug}
For any benign input $x$, and a mild transformation $W^{(g)}$, the resulting features satisfy
\begin{equation}
\| z^{(g)} - z \| \le \eta, \quad \| \ell^{(g)} - \ell \| \le \alpha(\eta),
\end{equation}
for small $\eta > 0$ and monotonically increasing $\alpha(\eta) \ll 1$.
\end{assumption}

\begin{assumption}[Recovery Test Evasion]\label{assump:recovery_fail}
Let $x^{\mathrm{adv}}=x+\delta$ be a successful adversarial input whose recovery errors do not exceed those of its clean counterpart,
\begin{equation}
\| z_i(x^{\mathrm{adv}}) - R^{(L \rightarrow i)}(z_L(x^{\mathrm{adv}})) \|^2 \lesssim \| z_i(x) - R^{(L \rightarrow i)}(z_L(x)) \|^2,
\end{equation}
while its final prediction differs from that of $x$,
\begin{equation}
\arg\max_c \ell_c(x^{\mathrm{adv}}) \ne \arg\max_c \ell_c(x).
\end{equation}
Suppose also that the learned augmentations produce feature changes of approximately the same magnitude for $x^{\mathrm{adv}}$ and $x$,
\begin{equation}
\| z^{(g), \text{adv}} - z^{\text{adv}} \| \approx \| z^{(g)} - z \|.
\end{equation}
\end{assumption}

\begin{theorem}[Discrete Consequence of Augmentation Instability]\label{thm:adv_logit_instab}
Let $\hat y=\arg\max_c\ell_c(x^{\mathrm{adv}})$, let $\mathbf p^{(g)}=\sigma(\ell(W^{(g)}x^{\mathrm{adv}}))$, and let $\hat y_g=\arg\max_c p_c^{(g)}$. If the learned transformation changes the predicted class, i.e., $\hat y\neq\hat y_g$, then the implemented LT prediction-deviation term obeys
\begin{equation}
\Delta\ell^{(g)}(x^{\mathrm{adv}})
=\|\mathbf o_{\hat y}-\mathbf p^{(g)}\|_2^2\geq\frac{1}{2}.
\end{equation}
\end{theorem}

(Proof: \autoref{pro:theorem2})

\begin{theorem}[Sufficient Condition for Amplified Logit Sensitivity]\label{thm:logit_amp}
For a fixed transformation $g$, define the positive numerator, denominator, and LT component
\begin{equation}
\begin{aligned}
N_g(u)&=\mathcal H(\sigma(\ell(u)))\Delta\ell^{(g)}(u)+\varepsilon,\\
D_g(u)&=\Delta z^{(g)}(u)+\varepsilon,\qquad
s_g(u)=\log N_g(u)-\log D_g(u).
\end{aligned}
\end{equation}
For a benign input $x$ and an adversarial input $x^{\mathrm{adv}}$, suppose
\begin{equation}
N_g(x^{\mathrm{adv}})>c_NN_g(x),\qquad
D_g(x^{\mathrm{adv}})\leq c_DD_g(x)
\end{equation}
for constants $c_N>c_D>0$. Then
\begin{equation}
s_g(x^{\mathrm{adv}})>s_g(x)+\log(c_N/c_D)>s_g(x).
\end{equation}
If the premises hold for every $g$, averaging gives $LT(x^{\mathrm{adv}})>LT(x)$.
\end{theorem}
(Proof: \autoref{pro:theorem3})

\subsection{Quantile-normalized RT + LT Provides Jointly Separable Score in RLT}

RLT fuses the two marginally normalized scores. Quantile normalization aligns their marginal scales while retaining any dependence between RT and LT; the separation result then uses an explicit score-margin condition.

\begin{assumption}[Quantile-normalized RT and LT]\label{assump:quantile}
Let $\mathcal{F}_{RT},\mathcal{F}_{LT}$ be continuous benign population CDFs for the population analysis and define
\begin{equation}
RT_{\text{norm}}(x) = \Phi^{-1}(\mathcal{F}_{RT}(RT(x))), \quad
LT_{\text{norm}}(x) = \Phi^{-1}(\mathcal{F}_{LT}(LT(x))),
\end{equation}
where $\Phi^{-1}$ is the standard normal quantile function. The implemented finite-sample scores instead use the clipped mid-rank empirical CDFs in Section~\ref{sec:calibration_protocol}.
\end{assumption}

\begin{assumption}[Adversarial Score Margin]\label{assump:margin}
Let $\tau$ be a calibrated RLT operating threshold. There exists $\gamma>0$ with $\gamma^2\geq\tau$ such that, for an adversarial input $x$, at least one normalized score satisfies
\begin{equation}
|RT_{\text{norm}}(x)| > \gamma \quad \text{or} \quad |LT_{\text{norm}}(x)| > \gamma.
\end{equation}
\end{assumption}

\begin{theorem}[Conditional RLT Separation]\label{thm:rlt}
Define the fused score:
\begin{equation}
RLT(x) = RT_{\text{norm}}(x)^2 + LT_{\text{norm}}(x)^2.
\end{equation}
For continuous benign population CDFs, each normalized benign marginal is standard normal, so $\mathbb E[RLT(x)]=2$ regardless of dependence. Under \autoref{assump:margin},
\begin{equation}
RLT(x^{\text{adv}})>\gamma^2\geq\tau,
\end{equation}
so the input is detected at threshold $\tau$. With finite empirical CDFs, the transformed distribution is discrete and the operating threshold is calibrated directly using Section~\ref{sec:calibration_protocol}.
\end{theorem}

(Proof: \autoref{pro:theorem4})

\subsection{Gradient Relationships under Adaptive Attacks}

An adaptive adversary knows the classifier and detector and explicitly optimizes misclassification and evasion. We first state the attack objective and then relate its component gradients to the direct measurements in \autoref{app:gradient_diagnostics}.

\subsubsection{Adaptive Attack Objective}

For an untargeted white-box adversary, the attack must both induce misclassification and evade the detector. We therefore minimize the following constrained objective:
\begin{equation}
\min_{\|\delta\| \leq \epsilon}
\left[-\mathcal{L}_{\mathrm{cls}}(x+\delta,y)
+ \beta_1 RT(x+\delta)+\beta_2 LT(x+\delta)\right],
\end{equation}
where the negative classification loss term is equivalent to maximizing cross-entropy for misclassification, and $\beta_1,\beta_2\ge 0$ trade off detector suppression. For a targeted attack toward $y_t$, the classification term becomes $\mathcal{L}_{\mathrm{cls}}(x+\delta,y_t)$. The fused-score attack used in \autoref{tab:bpda_pgd} sets $\beta_1,\beta_2$ through the quantile-normalized RLT score and sweeps the scalar detector weight $\lambda$.
While the classification term drives misclassification, the detector terms encourage lower RT and LT scores. In \autoref{tab:attack_measure_pairing}, each score is generally most vulnerable when attacked directly. \autoref{fig:lt_distributions} further shows that some attack-generated inputs have high LT scores even when no AFLS value $A_k$ exceeds its benign reference bound. Together, these observations explain the empirical complementarity of RT and LT.

\subsubsection{Potentially Competing Gradient Effects}\label{app:cge}

The implemented RT score in \autoref{eq:RT} depends on both the average magnitude of the reconstruction errors and their concentration across recovered layers. Lowering RT can therefore reduce the errors, distribute them more evenly across layers, or both. The implemented LT score in \autoref{eq:LT} averages
\begin{equation}
\log\!\left(\mathcal H(\sigma(\ell(x)))\Delta\ell^{(g)}(x)+\varepsilon\right)
-\log\!\left(\Delta z^{(g)}(x)+\varepsilon\right)
\end{equation}
over transformations $g$. Lowering LT can reduce the prediction deviation, increase the corresponding intermediate-feature change, or both. The measurements in \autoref{app:gradient_diagnostics} show that the misclassification and RT-suppression gradients are strongly opposed on successful PGD examples, whereas the relationship between the RT and LT gradients varies across samples. The benefit of fusion is therefore supported by their measured score complementarity rather than a consistently opposing RT--LT gradient direction.

\section{Proofs}
\label{app:assB}

\subsection{\autoref{thm:rec}}
\label{pro:theorem1}
\begin{proof}
Write $R=R^{(L\rightarrow i)}$. By the reverse triangle inequality,
\begin{equation}
\begin{aligned}
r_i(x')
&=\|z_i(x')-R(z_L(x'))\|_2\\
&\geq \|z_i(x')-z_i(x)\|_2
-\|z_i(x)-R(z_L(x))\|_2
-\|R(z_L(x'))-R(z_L(x))\|_2.
\end{aligned}
\end{equation}
Approximate invertibility gives $r_i(x)\leq\varepsilon$. Substituting the strict margin condition from \autoref{thm:rec} into the display above yields $r_i(x')>\varepsilon\geq r_i(x)$. Both quantities are nonnegative, so squaring preserves the strict inequality.
\end{proof}

\subsection{\autoref{thm:adv_logit_instab}}
\label{pro:theorem2}
\begin{proof}
Choose $j=\hat y_g\neq\hat y$. Then $p_j^{(g)}\geq p_{\hat y}^{(g)}$ and hence $p_{\hat y}^{(g)}\leq1/2$. Therefore
\begin{equation}
\|\mathbf o_{\hat y}-\mathbf p^{(g)}\|_2^2
=(1-p_{\hat y}^{(g)})^2+\sum_{c\neq\hat y}(p_c^{(g)})^2
\geq(1-p_{\hat y}^{(g)})^2+(p_j^{(g)})^2
\geq(1-p_{\hat y}^{(g)})^2+(p_{\hat y}^{(g)})^2
\geq\frac12.
\end{equation}
\end{proof}

\subsection{\autoref{thm:logit_amp}}
\label{pro:theorem3}
\begin{proof}
Using the two premises and positivity of $N_g,D_g$,
\begin{equation}
\frac{N_g(x^{\mathrm{adv}})}{D_g(x^{\mathrm{adv}})}
>
\frac{c_N}{c_D}\frac{N_g(x)}{D_g(x)}.
\end{equation}
Taking logarithms yields
$s_g(x^{\mathrm{adv}})>s_g(x)+\log(c_N/c_D)$, which is strictly larger than $s_g(x)$ because $c_N/c_D>1$. Averaging preserves the strict inequality when the premises hold for every transformation.
\end{proof}

\subsection{\autoref{thm:rlt}}
\label{pro:theorem4}
\begin{proof}
For continuous population CDFs, the probability integral transform makes each benign normalized marginal standard normal. Linearity of expectation, which does not require independence, gives
\begin{equation}
\mathbb E[RLT(x)]=\mathbb E[RT_{\mathrm{norm}}^2]+
\mathbb E[LT_{\mathrm{norm}}^2]=1+1=2.
\end{equation}
For an adversarial input satisfying \autoref{assump:margin}, at least one nonnegative squared component is strictly larger than $\gamma^2$; adding the other component preserves the strict lower bound, and $\gamma^2\geq\tau$ completes the threshold statement.
\end{proof}

\section{Ablation Study}
\label{app:assC}
All ablation studies were conducted using the CIFAR-10 dataset with ResNet110. We systematically vary the key hyperparameters and architectural choices to assess their impact on detection performance.

\subsection{Size of recovery module}

We evaluate the impact of varying the depth and dimensionality of the recovery modules on RT detection performance. \autoref{tab:recover_module_ablation} reports changes relative to depth 5 and dimensionality 512. Average AUC changes range from $-0.09$ to $+1.93$ points across depths and from $+0.13$ to $+1.47$ points across the smaller dimensions. The variation is modest overall, with the largest differences occurring for Square Attack.

\begin{table}[h]
\centering
\caption{RT AUC (\%) changes when varying the depth and dimensionality of the recovery modules, relative to depth 5 and dimensionality 512.}
\label{tab:recover_module_ablation}
\resizebox{0.7\textwidth}{!}{
\begin{tabular}{c|ccccc|c}
\toprule
\textbf{Depth} & \textbf{FGSM} & \textbf{PGD} & \textbf{CW} & \textbf{AutoAttack} & \textbf{Square} & \textbf{Avg.} \\
\midrule
2 & +0.55 & -2.03 & +0.16 & 0.00 & +10.96 & +1.93 \\
3 & +0.15 & +0.15 & +0.08 & 0.00 & +4.25 & +0.92 \\
4 & -0.01 & +0.01 & +0.04 & 0.00 & -0.50 & -0.09 \\
5 & 0.00 & 0.00 & 0.00 & 0.00 & 0.00 & 0.00 \\
\midrule
\textbf{Dimensionality} & \textbf{FGSM} & \textbf{PGD} & \textbf{CW} & \textbf{AutoAttack} & \textbf{Square} & \textbf{Avg.} \\
\midrule
64 & -0.50 & -0.10 & -0.40 & 0.00 & +8.34 & +1.47 \\
128 & -0.54 & -0.04 & -0.20 & 0.00 & +4.13 & +0.67 \\
256 & -0.18 & -0.18 & -0.03 & 0.00 & +0.65 & +0.13 \\
512 & 0.00 & 0.00 & 0.00 & 0.00 & 0.00 & 0.00 \\
\bottomrule
\end{tabular}
}
\end{table}

\subsection{Number of learnable augmentations $G$}

We investigate how the number of learnable transformation matrices ($G$) affects LT. Relative to $G=4$, the average AUC change ranges from $-0.01$ to $+0.12$ points, and every individual change is within $0.46$ points (\autoref{tab:augmentation_number_ablation}). Thus, one to four transformations perform similarly in this CIFAR-10/ResNet-110 experiment, supporting the use of a small $G$ in our evaluation.

\begin{table}[h]
\centering
\caption{LT test AUC (\%) differences when varying the number of augmentation matrices ($G$), compared to $G=4$.}
\label{tab:augmentation_number_ablation}
\resizebox{0.6\textwidth}{!}{
\begin{tabular}{c|ccccc|c}
\toprule
$\boldsymbol{G}$ & \textbf{FGSM} & \textbf{PGD} & \textbf{CW} & \textbf{AutoAttack} & \textbf{Square} & \textbf{Avg.} \\
\midrule
1 & +0.01 & -0.03 & -0.02 & -0.02 & +0.01 & -0.01 \\
2 & -0.04 & +0.08 & -0.14 & +0.37 & +0.03 & +0.06 \\
3 & 0.00 & +0.20 & -0.05 & +0.46 & +0.02 & +0.12 \\
4 & 0.00 & 0.00 & 0.00 & 0.00 & 0.00 & 0.00 \\
\bottomrule
\end{tabular}
}
\end{table}

\subsection{Choice of $k_{RT}$ and $k_{LT}$}

We examine how choosing the same value for $k_{RT}$ and $k_{LT}$ affects RLT relative to selecting them independently. In \autoref{tab:k_choice_ablation}, the average AUC decrease ranges from 0.83 to 1.33 points, while the largest decrease for an individual attack is 3.55 points. The average sensitivity is therefore modest in this experiment, although separate selection can improve particular attacks.

\begin{table}[h]
\centering
\caption{RLT test AUC (\%) differences for varying values of $k=k_{RT}=k_{LT}$ compared to optimal separate selection.}
\label{tab:k_choice_ablation}
\resizebox{0.7\textwidth}{!}{
\begin{tabular}{c|ccccc|c}
\toprule
\textbf{$\boldsymbol{k=k_{RT}=k_{LT}}$} & \textbf{FGSM} & \textbf{PGD} & \textbf{CW} & \textbf{AutoAttack} & \textbf{Square} & \textbf{Avg.} \\
\midrule
1 & -1.59 & -3.45 & -0.82 & -0.03 & -0.77 & -1.33 \\
10 & -1.37 & -3.55 & -0.49 & -0.04 & -0.86 & -1.26 \\
15 & -1.03 & -3.41 & -0.35 & -0.04 & -1.02 & -1.17 \\
20 & -1.25 & -3.16 & -0.33 & -0.04 & -0.63 & -1.08 \\
25 & -1.37 & -2.51 & -0.37 & -0.03 & -1.05 & -1.06 \\
30 & -1.67 & -1.33 & -0.39 & -0.01 & -0.77 & -0.83 \\
$k_{RT}\neq k_{LT}$ (Optimal) & 0.00 & 0.00 & 0.00 & 0.00 & 0.00 & 0.00 \\
\bottomrule
\end{tabular}
}
\end{table}

\subsection{Single-Objective Adaptive Attacks}

We first ablate cases where the attacker and defender use a single detection objective, RT or LT. As shown in \autoref{tab:single_adaptive_attack}, removing either objective reduces system RA relative to the jointly attacked and measured defense in this Orthogonal-PGD protocol, demonstrating the empirical complementarity of RT and LT.

\begin{table}[h]
\centering
\caption{Single-objective Orthogonal-PGD adaptive attack at $L_\infty=8/255$ using CIFAR-10 and ResNet-110.}
\label{tab:single_adaptive_attack}
\resizebox{0.5\textwidth}{!}{
\begin{tabular}{lcc}
\toprule
\textbf{Removed Objective} & \textbf{RA@FPR5\%} & \textbf{RA@FPR50\%} \\
\midrule
\textbf{None} & \textbf{33.70\%} & \textbf{80.77\%} \\
\textbf{RT} & 17.11\% & 55.30\% \\
\textbf{LT} & 17.20\% & 56.17\% \\
\bottomrule
\end{tabular}
}
\end{table}

We next mismatch the attack objective and measured defense score. In \autoref{tab:attack_measure_pairing}, attacking RLT gives the lowest average system RA across the three measurement scores (64.85\%), while the diagonal entries show that each score is generally most vulnerable when attacked directly. These results motivate attacking the deployed fused score rather than relying on a transfer attack against one component.

\begin{table}[h]
\centering
\caption{System RA at FPR 50\% when the end-to-end PGD attack objective and evaluated detector score differ (an $L_\infty$ budget of $\epsilon=8/255$, ResNet-110, CIFAR-10).}
\label{tab:attack_measure_pairing}
\resizebox{0.7\textwidth}{!}{
\begin{tabular}{lccc}
\toprule
\textbf{Evaluated score $\downarrow$ / Attack objective $\rightarrow$} & \textbf{RLT} & \textbf{RT} & \textbf{LT} \\
\midrule
\textbf{RLT} & $\mathbf{68.53_{\pm2.21}}$ & $72.06_{\pm4.05}$ & $73.05_{\pm2.99}$ \\
\textbf{RT} & $78.75_{\pm1.53}$ & $\mathbf{75.56_{\pm1.30}}$ & $81.58_{\pm1.73}$ \\
\textbf{LT} & $47.28_{\pm4.43}$ & $64.49_{\pm8.12}$ & $\mathbf{46.48_{\pm3.67}}$ \\
\midrule
\textbf{Average} & \textbf{64.85} & 70.70 & 67.04 \\
\bottomrule
\end{tabular}
}
\end{table}

The mismatch results motivate a direct analysis of the corresponding input-gradient directions, which we provide next.

\subsection{Direct Gradient-Alignment Diagnostics}\label{app:gradient_diagnostics}

We measure input-gradient cosine similarities on CIFAR-10/ResNet-110 for $1{,}000$ inputs correctly classified before perturbation and $1{,}000$ successful 50-step PGD examples ($\epsilon=8/255$). All vectors are gradients of terms in the minimized adaptive objective: $g_{\mathrm{cls}}=\nabla_x[-\mathcal L_{\mathrm{cls}}]$, $g_{RT}=\nabla_x[RT_{\mathrm{norm}}^2]$, and $g_{LT}=\nabla_x[LT_{\mathrm{norm}}^2]$. The forward pass uses exact empirical-CDF values; as in the adaptive evaluation, the CDF backward pass uses an affine standardized BPDA approximation.

\begin{table}[h]
\centering
\caption{Input-gradient cosine similarities. We report the median, interquartile range (IQR), and percentage of samples with negative cosine.}
\label{tab:gradient_alignment}
\begingroup
\resizebox{0.85\textwidth}{!}{
\begin{tabular}{llccc}
\toprule
\textbf{Input setting} & \textbf{Gradient pair} & \textbf{Median cosine} & \textbf{IQR} & \textbf{Negative cosine (\%)} \\
\midrule
Clean & $(g_{\mathrm{cls}},g_{RT})$ & $-0.031$ & $[-0.236,\;0.191]$ & 54.2 \\
Clean & $(g_{\mathrm{cls}},g_{LT})$ & $-0.070$ & $[-0.614,\;0.082]$ & 61.3 \\
Clean & $(g_{RT},g_{LT})$ & $\phantom{-}0.021$ & $[-0.102,\;0.143]$ & 45.2 \\
\midrule
Successful PGD & $(g_{\mathrm{cls}},g_{RT})$ & $-0.762$ & $[-0.846,\;-0.651]$ & 99.8 \\
Successful PGD & $(g_{\mathrm{cls}},g_{LT})$ & $\phantom{-}0.088$ & $[-0.171,\;0.194]$ & 40.7 \\
Successful PGD & $(g_{RT},g_{LT})$ & $-0.053$ & $[-0.160,\;0.130]$ & 58.1 \\
\bottomrule
\end{tabular}
}
\endgroup
\end{table}

On successful PGD examples, the misclassification and RT-suppression gradients are strongly opposed: the median cosine is $-0.762$, the IQR is entirely negative, and 99.8\% of samples have negative cosine. In contrast, the RT--LT median is $-0.053$ with an IQR spanning zero, showing that the relative direction of the two detector gradients varies across samples.

\subsection{Adaptive Metrics and Sensitivity}\label{app:adaptive_scope}

Among inputs correctly classified before perturbation, let classification ASR be the fraction misclassified after attack, conditional evasion be the fraction of successful attacks with detector score at or below the benign threshold, and joint success rate (JSR) be the fraction satisfying both events. The reported system robust accuracy is $RA=1-\mathrm{JSR}$, so it credits either a correct prediction or detection of a misclassified input. Reporting only detector evasion or only classification ASR would not characterize the defended system.

The end-to-end BPDA tables sweep $\lambda\in\{0.25,0.50,1.00\}$ with five repeated runs. At the low-FPR operating point, standalone RLT RA ranges from $20.72\%$ to $23.39\%$ at FPR 5\%, showing that the tested white-box objective substantially stresses the detector alone. With the ATC, RA ranges from $93.32\%$ to $93.57\%$ at the same operating point. Orthogonal-PGD provides a complementary adaptive objective without scalar loss balancing, while SimBA checks the classifier prediction path without first-order gradients. Together, these experiments examine weighted and orthogonal white-box objectives and a gradient-free classifier-output attack; further adaptive strategies are discussed in the limitations.

\subsection{Ablation on perturbation budget sensitivity of each testing measure}

\begin{figure}[h]
    \centering
    \includegraphics[width=1.0\linewidth]{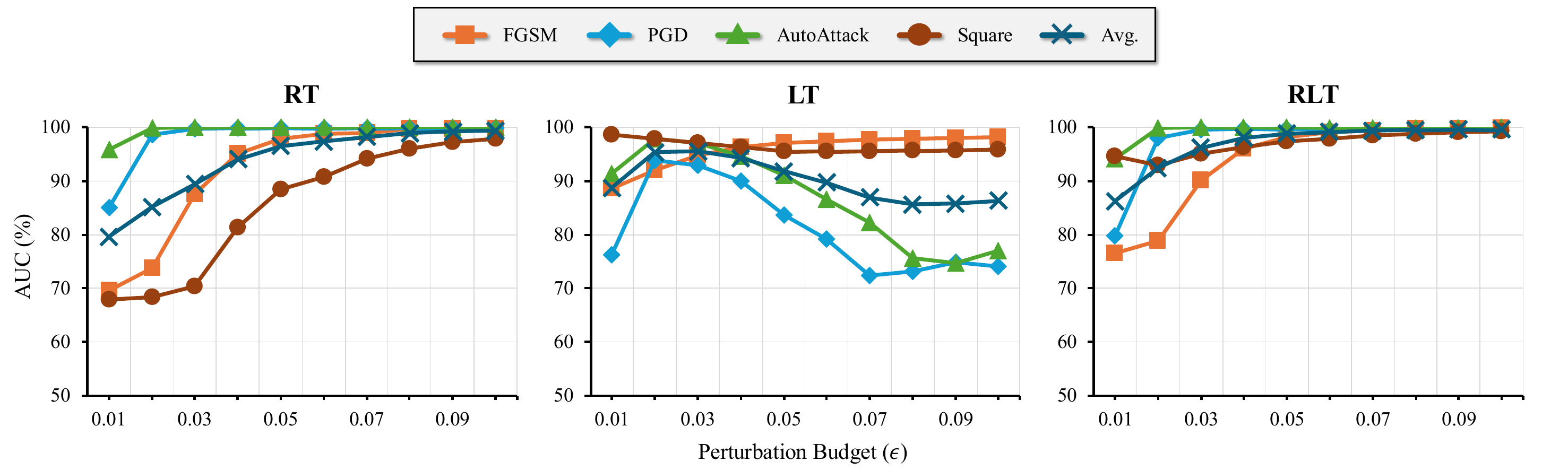}
    \caption{
       Ablation study showing AUC (\%) of RT, LT, and RLT under varying perturbation budgets ($\epsilon \in \{0.01, 0.02, \cdots, 0.1\}$) across multiple standard attack types (FGSM, PGD, AutoAttack, Square). Evaluated on CIFAR-10 using a ResNet-110 classifier and detection models trained with fewer epochs for efficiency.
    }
    \label{fig:AUC_by_pert}
\end{figure}

To analyze the contribution of each detection component across adversarial strengths, we conduct an ablation study varying the perturbation budget $\epsilon$ for common standard (non-adaptive) attacks, including FGSM, PGD, AutoAttack, and Square. We report AUC detection performance for RT, LT, and RLT. All experiments are performed on the CIFAR-10 dataset using a pretrained ResNet-110 classifier. Detection models are trained with a reduced number of epochs to simulate lightweight deployment.

In \autoref{fig:AUC_by_pert}, RT and RLT generally increase with perturbation budget, whereas LT often decreases; the detailed pattern depends on the attack.

RT and LT respond differently as the budget changes: in several curves RT improves while LT declines, and their fusion varies more smoothly than either component. These score trends provide further evidence that RT and LT capture complementary behavior.

These score trends are compatible with AFLS for the tested budgets, while the direct comparison of clean and perturbed representations in \autoref{app:additional_afls_validation} is the evaluation of that hypothesis.

\subsection{Contribution of each term in testing measures}

\begin{figure}[h]
    \centering
    \begin{overpic}[width=1.0\linewidth]{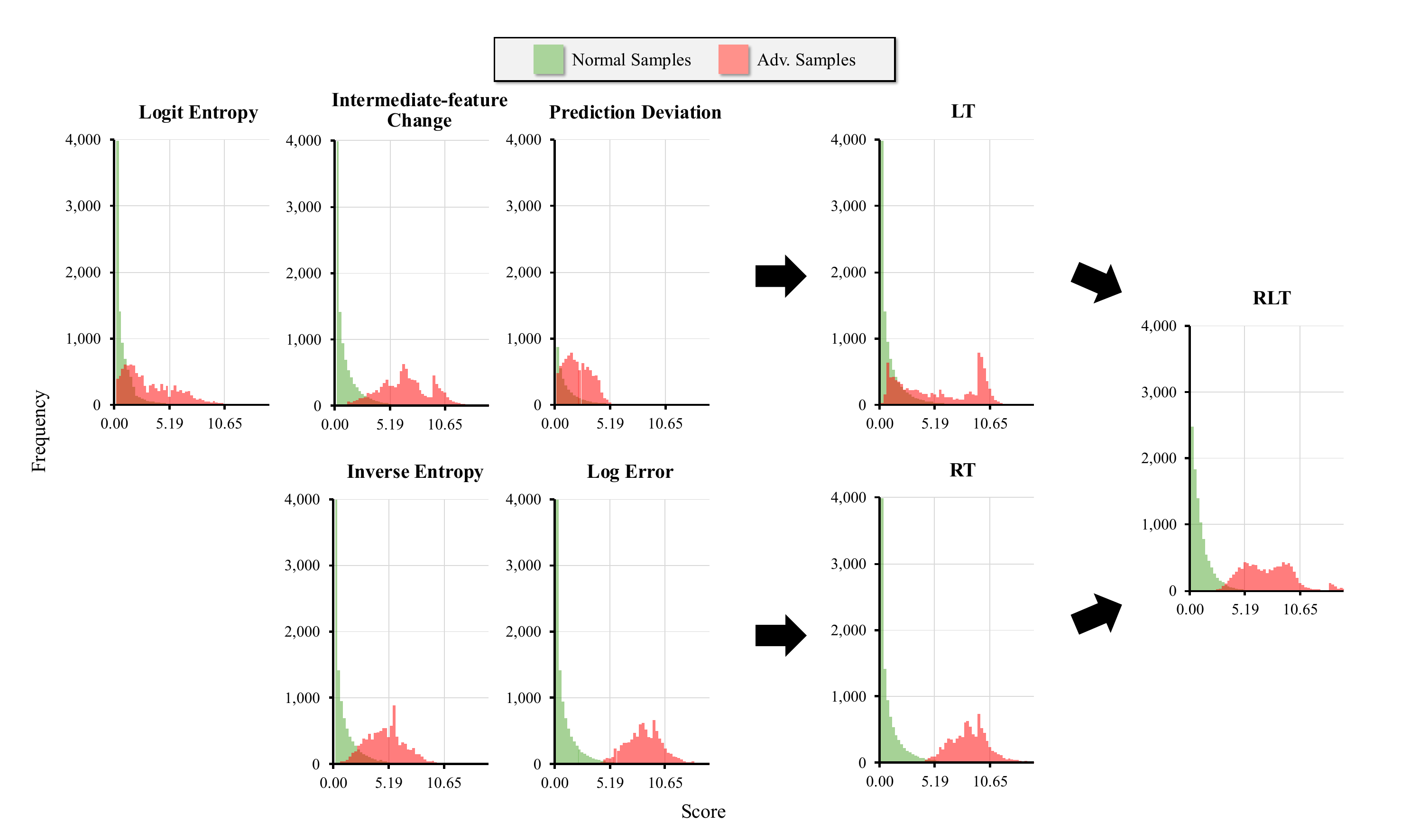}
    \end{overpic}
    \caption{
       Empirical distributions of individual terms used in our detection metrics on CIFAR-10/ResNet-110 under FGSM ($\epsilon = 0.05$). Scores are squared and normalized using quantile normalization fitted on the designated benign CDF-fitting samples.
    }
    \label{fig:term_distribution}
\end{figure}

\begin{table}[h]
\centering
\caption{Contribution of each term in LT and RT. AUC (\%) is reported on CIFAR-10/ResNet-110. The last column is the average change relative to the full score.}
\label{tab:ablation_contributions}
\resizebox{1\textwidth}{!}{
\begin{tabular}{lcccccc|c}
\toprule
\textbf{Removed Term} & \textbf{FGSM} & \textbf{PGD} & \textbf{CW} & \textbf{AutoAttack} & \textbf{Square} & \textbf{Average} & \textbf{Avg. Change} \\
\midrule
\multicolumn{8}{l}{\textbf{Logit-layer Testing (LT)}} \\
None (Full LT)            & 97.50 & 98.61 & 97.08 & 99.60 & 97.47 & 98.05 & – \\
\quad – Prediction Deviation          & 99.51 & 94.66 & 82.97 & 96.42 & 95.91 & 93.89 & -4.16 \\
\quad – Logit Entropy                & 97.51 & 59.37 & 89.76 & 57.14 & 97.20 & 80.19 & -17.86 \\
\quad – Intermediate-feature Change  & 94.62 & 65.49 & 99.29 & 50.01 & 97.29 & 81.34 & -16.71 \\
\midrule
\multicolumn{8}{l}{\textbf{Recovery Testing (RT)}} \\
None (Full RT)            & 99.93 & 96.89 & 99.90 & 99.99 & 85.38 & 96.42 & – \\
\quad – Inverse Entropy             & 98.98 & 94.35 & 99.09 & 99.95 & 84.52 & 95.38 & -1.04 \\
\quad – Log Error                   & 80.34 & 97.09 & 88.95 & 99.95 & 65.68 & 86.40 & -10.02 \\
\bottomrule
\end{tabular}
}
\end{table}

To better understand the role of each component in our proposed detection metrics, we perform an ablation study by individually removing key terms from the Logit-layer Testing and Recovery Testing scores. We evaluate the resulting change in detection performance using AUC scores under five adversarial attack types--FGSM, PGD, CW, AutoAttack, and Square--on CIFAR-10 with a ResNet-110 classifier. \autoref{tab:ablation_contributions} summarizes the results and reports the signed average AUC change relative to each full score.

For LT, removing output entropy lowers average AUC by 17.86 points, removing the intermediate-feature-change term lowers it by 16.71 points, and removing prediction deviation lowers it by 4.16 points. Some individual attacks are exceptions: removing prediction deviation improves FGSM, and removing the intermediate-feature-change term improves CW. Thus, all three terms improve the average result, while their contributions depend on the attack.

For RT, removing the log-error magnitude lowers average AUC by 10.02 points, whereas removing inverse entropy lowers it by 1.04 points. PGD is an exception for the log-error ablation, improving by 0.20 points. The magnitude term is therefore the larger average contributor in this setting, and inverse entropy provides a smaller average gain.

These ablations indicate that RT and LT use distinct cues and that all tested terms contribute to their average performance.

\autoref{fig:term_distribution} complements \autoref{tab:ablation_contributions} for its displayed FGSM setting: individual terms show different degrees of separation, while the combined RT, LT, and RLT scores show clearer separation in this diagnostic.

\subsection{Evaluation under an $L_2$-Norm Attack}
\label{app:l2_norm_attacks}

We additionally evaluate an $L_2$-PGD configuration to complement the main $L_\infty$ evaluation.

In the tested budget sequence, $\epsilon=0.6$ is the first value that reduces the base classifier's robust accuracy to 0.00\% (\autoref{tab:l2_budget_finding}), and we use this budget for the detector comparison.

\begin{table}[h]
\centering
\caption{Base-classifier RA (\%) across the tested $L_2$-PGD budgets on CIFAR-10.}
\label{tab:l2_budget_finding}
\resizebox{0.12\textwidth}{!}{%
\begin{tabular}{cc}
\toprule
$\boldsymbol{\epsilon}$ & \textbf{RA} \\
\midrule
0.1 & 39.97 \\
0.2 & 6.21 \\
0.3 & 0.74 \\
0.4 & 0.10 \\
0.5 & 0.01 \\
0.6 & 0.00 \\
\bottomrule
\end{tabular}
}
\end{table}

At this selected budget, RLT achieves 98.05\% AUC under $L_2$-PGD, compared with 99.47\% under the reported $L_\infty$-PGD configuration.

\begin{table}[h]
\centering
\caption{AUC (\%) scores comparing performance against $L_\infty$ and $L_2$ PGD attacks on CIFAR-10.}
\label{tab:l2_vs_linf_auc}
\resizebox{0.35\textwidth}{!}{%
\begin{tabular}{lcccc}
\toprule
\textbf{Norm} & \textbf{RA} & \textbf{RLT} & \textbf{RT} & \textbf{LT} \\
\midrule
$L_\infty$ & 0.00 & \textbf{99.47} & \textbf{99.49} & 96.65 \\
$L_2$ & 0.00 & 98.05 & 95.05 & \textbf{97.16} \\
\bottomrule
\end{tabular}
}
\end{table}

\section{Evaluation across CNN Architectures}
\label{sec:arch_eval}

\begin{table*}[h]
    \centering
    \caption{AUC scores (\%) on CIFAR-100 under three white‑box attacks
    (FGSM, $\ell_\infty$‑PGD, and AutoAttack) with the same perturbation budget as the previous evaluations.}
    \label{tab:cifar100_robustness}
    
    \setlength{\tabcolsep}{6pt}
    \resizebox{0.75\textwidth}{!}{
    \begin{tabular}{lcccccccc}
        \toprule
        & \multicolumn{4}{c}{\textbf{ResNet‑18}} & \multicolumn{4}{c}{\textbf{MobileNet‑V2 x0.5}}\\
        \cmidrule(lr){2-5}\cmidrule(lr){6-9}
        \textbf{Variant} & FGSM & PGD & AutoAttack & \textbf{Avg.}
                        & FGSM & PGD & AutoAttack & \textbf{Avg.}\\
        \midrule
        RT  & 87.83 & 99.21 & 99.98 & 95.67
            & 99.02 & 99.31 & 99.99 & 99.44\\
        LT  & 97.83 & 97.53 & 99.47 & 98.28
            & 94.73 & 87.50 & 91.74 & 91.33\\
        RLT & 95.61 & 98.92 & 99.97 & 98.17
            & 98.73 & 98.14 & 99.98 & 98.95\\
        \midrule\midrule
        & \multicolumn{4}{c}{\textbf{ShuffleNet‑V2 x0.5}} & \multicolumn{4}{c}{\textbf{RepVGG a0}}\\
        \cmidrule(lr){2-5}\cmidrule(lr){6-9}
        RT  & 98.38 & 97.35 & 99.97 & 98.57
            & 91.77 & 98.94 & 99.99 & 96.90\\
        LT  & 94.82 & 90.34 & 96.81 & 93.99
            & 90.38 & 96.57 & 99.26 & 95.40\\
        RLT & 98.16 & 96.17 & 99.94 & 98.09
            & 93.49 & 98.46 & 99.98 & 97.31\\
        \bottomrule
    \end{tabular}
    }
\end{table*}

To assess transfer across CNN families, we evaluate RT, LT, and RLT on CIFAR-100 using ResNet-18, MobileNet-V2 x0.5, ShuffleNet-V2 x0.5, and RepVGG-a0. In \autoref{tab:cifar100_robustness}, RLT has average AUC above 97\% for each of these four targets under FGSM, PGD, and AutoAttack.

Together with the ResNet and DenseNet main experiments, these results demonstrate transfer across distinct CNN targets, including lightweight and re-parameterized convolutional families.

The target classifier remains fixed when applying RLT; deployment entails selecting its feature taps, fitting the auxiliary modules, and calibrating the detector on representative benign data.

\section{Evaluation on a Supervised ViT-B/16}
\label{app:vit}

To test one target beyond CNNs, we apply the detector to a supervised ViT-B/16 on CIFAR-10 and CIFAR-100. The results are presented in \autoref{tab:vit_results}.

We adapted our framework to the ViT-B/16 architecture by extracting features from its constituent layers. The output of the final (12th) encoder block was designated as the source embedding ($z_L$), from which all prior layer representations were reconstructed. For the initial \texttt{conv\_proj} layer, patch embeddings were spatially averaged into 14 vertical strips, with a dedicated recovery module trained for each. For each of the 11 preceding Transformer encoder blocks, the $197 \times 768$ token embedding matrices were partitioned into 6 non-overlapping feature chunks. Each chunk was then aggregated to create a feature vector for its corresponding recovery module. For LT, we used $G=4$ augmentations. In total, this configuration uses 80 recovery modules, whose combined parameter count is 0.337$\times$ that of the target ViT model.

RLT exceeds LID for all three attacks on CIFAR-10; on CIFAR-100 it exceeds LID for PGD and CW and is lower for FGSM. These results demonstrate the feasibility of adapting RLT to a supervised Transformer target.

As with the CNN targets, the ViT adaptation uses model-specific feature selection, benign fitting, and calibration while leaving the target classifier unchanged.

\begin{table}[h]
\centering
\caption{AUC (\%) scores for ViT-B/16 on CIFAR-10 and CIFAR-100 against various attacks ($\epsilon = 0.03$).}
\label{tab:vit_results}
\resizebox{0.45\textwidth}{!}{
\begin{tabular}{l l ccc}
\toprule
\textbf{Dataset} & \textbf{Method} & \textbf{FGSM} & \textbf{PGD} & \textbf{CW} \\
\midrule
\multirow{2}{*}{CIFAR-10} 
 & LID & 92.65 & 82.89 & 67.90 \\
 & \textbf{Ours (RLT)} & \textbf{95.14} & \textbf{90.68} & \textbf{99.99} \\
\midrule
\multirow{2}{*}{CIFAR-100} 
 & LID & \textbf{91.05} & 81.28 & 74.37 \\
 & \textbf{Ours (RLT)} & 88.18 & \textbf{84.54} & \textbf{99.99} \\
\bottomrule
\end{tabular}
}
\end{table}

\section{Detailed Computational Cost Analysis}
\label{app:overhead}

To substantiate our claim of low computational overhead, we provide a detailed analysis comparing our framework to two common detection paradigms: SSL-based and reference-set-based detectors. The analysis uses the target models detailed in \autoref{tab:overhead_models}.

First, SSL-based detectors require large, pre-trained models, which introduce a substantial and fixed overhead. As shown in \autoref{tab:overhead_ssl}, this cost is particularly prohibitive for lightweight target models, with overheads ranging from 3.21$\times$ to over 200$\times$ the parameters of the base classifier.

Second, reference-set detectors incur significant memory costs for storing embeddings and, in the case of graph-based methods, adjacency matrices. As detailed in \autoref{tab:overhead_refset}, this overhead scales with the size of the reference set and can be prohibitively large, especially for graph-based approaches.

The framework provides a tunable overhead through recovery-module depth and width (\autoref{tab:overhead_ours}). On ResNet-110, the evaluated 0.24$\times$ configuration achieves 99.57\% AUC, less than 0.3 percentage points below the larger tested configuration, illustrating the measured cost--performance trade-off.

\begin{table}[h]
\centering
\caption{Target models used in our experiments, along with their parameter counts and the number of feature blocks (layers) used for applying RT.}
\label{tab:overhead_models}
\resizebox{0.5\textwidth}{!}{
\begin{tabular}{lrr}
\toprule
\textbf{Target Model} & \textbf{\# Parameters} & \textbf{\# Feature Blocks} \\
\midrule
MobileNet & 815,780 & 19 \\
ShuffleNet & 1,356,104 & 18 \\
ResNet110 & 1,730,714 & 56 \\
RepVGG & 7,956,164 & 23 \\
\bottomrule
\end{tabular}
}
\end{table}

\begin{table}[h]
\centering
\caption{Evaluation of SSL-based approaches’ minimum overhead ($\times$) introduced by pre-trained SSL models relative to each target model.}
\label{tab:overhead_ssl}
\resizebox{0.7\textwidth}{!}{
\begin{tabular}{lrrrrr}
\toprule
\textbf{SSL Models} & \textbf{\# Params} & \textbf{MobileNet} & \textbf{ShuffleNet} & \textbf{ResNet110} & \textbf{RepVGG} \\
\midrule
BYOL & 25,557,032 & 31.33 & 18.85 & 14.77 & 3.21 \\
SimSiam & 38,201,408 & 46.83 & 28.17 & 22.07 & 4.80 \\
MoCo v3 (ViT) & 215,678,464 & 264.38 & 159.04 & 124.62 & 27.11 \\
\bottomrule
\end{tabular}
}
\end{table}

\begin{table}[h]
\centering
\caption{Minimum overhead ($\times$) from detection approaches using a reference set with a 1024-dimensional embedding space. "Graph Structure" indicates whether the approach constructs an adjacency matrix.}
\label{tab:overhead_refset}
\resizebox{\textwidth}{!}{
\begin{tabular}{lrrrrrr}
\toprule
\textbf{Graph Structure} & \textbf{Reference Set Size} & \textbf{\# Params} & \textbf{MobileNet} & \textbf{ShuffleNet} & \textbf{ResNet110} & \textbf{RepVGG} \\
\midrule
No & 1,000 & 1,024,000+ & 1.26+ & 0.76+ & 0.59+ & 0.13+ \\
 & 5,000 & 5,120,000+ & 6.28+ & 3.78+ & 2.96+ & 0.64+ \\
 & 40,000 & 40,960,000+ & 50.21+ & 30.20+ & 23.67+ & 5.15+ \\
\midrule
Yes & 1,000 & 2,024,000+ & 2.48+ & 1.49+ & 1.17+ & 0.25+ \\
 & 5,000 & 30,120,000+ & 36.92+ & 22.21+ & 17.40+ & 3.79+ \\
 & 40,000 & 1,640,960,000+ & 2011.52+ & 1210.05+ & 948.14+ & 206.25+ \\
\bottomrule
\end{tabular}
}
\end{table}

\begin{table}[h]
\centering
\caption{Parameter overhead ($\times$) and AUC (\%) on CIFAR-100 under $\ell_\infty$-PGD ($\epsilon=0.02$) for recovery-module depth/width choices. ``AUC difference'' is the listed AUC minus that of the largest configuration for the same target; negative values indicate lower AUC and positive values higher AUC.}
\label{tab:overhead_ours}
\resizebox{0.9\textwidth}{!}{
\begin{tabular}{lrrrrrr}
\toprule
\textbf{Target Model} & \textbf{Depth} & \textbf{Width} & \textbf{Detector's \# Params} & \textbf{Overhead} & \textbf{AUC} & \textbf{AUC Difference} \\
\midrule
\multirow{6}{*}{MobileNet} & 2 & 64 & 1,540,744 & 1.89 & 97.37 & 0.0000 \\
 & 2 & 32 & 776,904 & 0.95 & 97.20 & -0.1665 \\
 & 2 & 16 & 394,984 & 0.48 & 96.07 & -1.3025 \\
 & 2 & 8 & 204,024 & 0.25 & 95.12 & -2.2492 \\
 & 2 & 4 & 108,544 & 0.13 & 89.27 & -8.0933 \\
 & 2 & 2 & 60,804 & 0.07 & 87.66 & -9.7059 \\
\midrule
\multirow{9}{*}{ShuffleNet} & 2 & 128 & 2,788,840 & 2.06 & 96.07 & 0.0000 \\
 & 4 & 64 & 1,548,456 & 1.14 & 95.46 & -0.6154 \\
 & 2 & 64 & 1,402,664 & 1.03 & 95.96 & -0.1152 \\
 & 4 & 32 & 747,656 & 0.55 & 95.12 & -0.9522 \\
 & 2 & 32 & 709,576 & 0.52 & 95.48 & -0.5946 \\
 & 4 & 16 & 373,368 & 0.28 & 93.88 & -2.1923 \\
 & 2 & 16 & 363,032 & 0.27 & 95.25 & -0.8218 \\
 & 4 & 8 & 192,752 & 0.14 & 93.02 & -3.0472 \\
 & 2 & 8 & 189,760 & 0.14 & 94.56 & -1.5079 \\
\midrule
\multirow{8}{*}{ResNet110} & 3 & 256 & 5,160,992 & 2.98 & 99.84 & 0.0000 \\
 & 2 & 256 & 1,514,272 & 0.87 & 99.58 & -0.2640 \\
 & 3 & 64 & 652,448 & 0.38 & 99.71 & -0.1259 \\
 & 2 & 64 & 416,608 & 0.24 & 99.57 & -0.2717 \\
 & 3 & 16 & 158,912 & 0.09 & 99.51 & -0.3313 \\
 & 2 & 16 & 142,192 & 0.08 & 99.41 & -0.4252 \\
 & 3 & 4 & 75,128 & 0.04 & 99.62 & -0.2180 \\
 & 2 & 4 & 73,588 & 0.04 & 99.63 & -0.2119 \\
\midrule
\multirow{10}{*}{RepVGG} & 4 & 256 & 11,310,992 & 1.42 & 98.45 & 0.0000 \\
 & 3 & 256 & 9,852,304 & 1.24 & 98.47 & 0.0144 \\
 & 4 & 128 & 4,942,992 & 0.62 & 98.38 & -0.0721 \\
 & 3 & 128 & 4,574,096 & 0.57 & 98.41 & -0.0386 \\
 & 4 & 64 & 2,299,664 & 0.29 & 98.34 & -0.1092 \\
 & 3 & 64 & 2,205,328 & 0.28 & 98.36 & -0.0969 \\
 & 4 & 32 & 1,113,168 & 0.14 & 98.23 & -0.2185 \\
 & 3 & 32 & 1,088,528 & 0.14 & 98.32 & -0.1341 \\
 & 4 & 16 & 553,712 & 0.07 & 98.01 & -0.4442 \\
 & 3 & 16 & 547,024 & 0.07 & 98.11 & -0.3468 \\
\bottomrule
\end{tabular}
}
\end{table}

\section{Detailed System-Level Analysis}
\label{app:system_level_analysis}

In this section, we provide a detailed theoretical and empirical analysis of the system-level performance when our detector is applied to a standard classifier. We formally define the evaluation metrics, introduce a framework for establishing lower bounds on system accuracy under these metric definitions, and demonstrate how this framework can be used to select detection thresholds in practical scenarios.

\subsection{Definitions of Metrics}
To formally analyze performance, we define metrics for both the base classifier and the combined classifier-detector system. Let $C(x)$ be the classifier's prediction for an input $x$, $y$ be its true label, and $D(x)$ be our detector's output, where $D(x)=1$ signifies an adversarial detection.
\begin{itemize}
    \item \textbf{Classifier Clean Accuracy ($CA_{cls}$)}: The accuracy of the base classifier on benign samples without any detector. $CA_{cls} = \mathbb{P}(C(x_{\text{clean}}) = y)$.
    \item \textbf{Classifier Robust Accuracy ($RA_{cls}$)}: The accuracy of the base classifier on adversarial samples without any detector. $RA_{cls} = \mathbb{P}(C(x_{\text{adv}}) = y)$.
    \item \textbf{System Clean Accuracy ($CA_{sys}$)}: The accuracy of the combined system on benign samples. A benign sample is handled correctly only if it is both correctly classified and not flagged by the detector.
    \begin{equation}CA_{sys} = \mathbb{E}[\mathbb{I}(C(x_{\text{clean}}) = y \land D(x_{\text{clean}}) = 0)]\end{equation}
    \item \textbf{System Robust Accuracy ($RA_{sys}$)}: The accuracy of the combined system on adversarial samples. An adversarial sample is successfully defended if it is either detected or correctly classified despite the attack.
    \begin{equation}RA_{sys} = \mathbb{E}[\mathbb{I}(C(x_{\text{adv}}) = y \lor D(x_{\text{adv}}) = 1)]\end{equation}
    \item \textbf{Overall System Accuracy ($A_{sys}$)}: The expected accuracy of the system given a probability $p$ that an input is adversarial.
    \begin{equation}A_{sys} = (1-p) \cdot CA_{sys} + p \cdot RA_{sys}\end{equation}
    \item \textbf{False Positive Rate (FPR)}: The fraction of benign samples incorrectly flagged as adversarial. $FPR = \mathbb{P}(D(x_{\text{clean}}) = 1)$.
    \item \textbf{True Positive Rate (TPR)}: The fraction of adversarial samples correctly flagged as adversarial. $TPR = \mathbb{P}(D(x_{\text{adv}}) = 1)$.
\end{itemize}

\subsection{Performance Analysis Across Operating Points}
While AUC provides an aggregate measure, evaluating performance at fixed operating points is critical for understanding the practical trade-off between clean accuracy and robustness.

First, we establish the baseline performance of the target ResNet-110 classifier \textit{without} our defense in \autoref{tab:baseline_accuracy}. The results show that while the model achieves high clean accuracy, its robustness is completely compromised by strong attacks like PGD and AutoAttack, with the $RA_{cls}$ dropping to 0.00\%.

\begin{table}[h]
\centering
\caption{Baseline CA and RA (\%) of the undefended ResNet-110 on CIFAR-10 ($\epsilon=8/255$).}
\label{tab:baseline_accuracy}
\resizebox{0.25\textwidth}{!}{
\begin{tabular}{lcc}
\toprule
\textbf{Attack} & \textbf{$CA_{cls}$} & \textbf{$RA_{cls}$} \\
\midrule
FGSM & 92.49 & 25.75 \\
PGD & 92.49 & 0.00 \\
CW & 92.49 & 47.17 \\
AutoAttack & 92.49 & 0.00 \\
\bottomrule
\end{tabular}
}
\end{table}

With the RLT detector active, the operating-point results are reported in \autoref{tab:operating_points}. At 5\% FPR, the system has $CA_{sys}=88.70\%$ and $RA_{sys}=99.27\%$ under PGD, compared with $RA_{cls}=0\%$ for the classifier alone. Across the reported thresholds, increasing FPR raises TPR and $RA_{sys}$ while lowering $CA_{sys}$.

\begin{table}[h]
\centering
\caption{Measured TPR, System Clean Accuracy ($CA_{sys}$), and System Robust Accuracy ($RA_{sys}$) (\%) at varying FPRs for the defended system on CIFAR-10 ($\epsilon=8/255$).}
\label{tab:operating_points}
\resizebox{0.7\textwidth}{!}{
\begin{tabular}{lcccccc}
\toprule
\textbf{Attack} & \textbf{Metric} & \textbf{@FPR5\%} & \textbf{@FPR10\%} & \textbf{@FPR25\%} & \textbf{@FPR50\%} \\
\midrule
\multirow{3}{*}{FGSM} & TPR & 84.84 & 94.13 & 99.21 & \textbf{99.94} \\
& $CA_{sys}$ & \textbf{88.71} & 84.52 & 71.60 & 48.80 \\
& $RA_{sys}$ & 90.80 & 96.57 & 99.58 & \textbf{99.96} \\
\midrule
\multirow{3}{*}{PGD} & TPR & 99.27 & 99.68 & 99.92 & \textbf{99.95} \\
& $CA_{sys}$ & \textbf{88.70} & 84.73 & 71.71 & 48.70 \\
& $RA_{sys}$ & 99.27 & 99.68 & 99.92 & \textbf{99.95} \\
\midrule
\multirow{3}{*}{CW} & TPR & 90.46 & 96.48 & 99.48 & \textbf{99.97} \\
& $CA_{sys}$ & \textbf{88.63} & 84.57 & 71.79 & 48.77 \\
& $RA_{sys}$ & 96.14 & 98.79 & 99.91 & \textbf{100.00} \\
\midrule
\multirow{3}{*}{AutoAttack} & TPR & 88.86 & 89.90 & 91.88 & \textbf{93.87} \\
& $CA_{sys}$ & \textbf{88.70} & 84.73 & 71.71 & 48.70 \\
& $RA_{sys}$ & 88.86 & 89.90 & 91.88 & \textbf{93.87} \\
\bottomrule
\end{tabular}
}
\end{table}

\subsection{Plug-in Robustness Gains with Adversarially Trained Classifiers (ATC)}
\label{app:atc_gains}

A key advantage of our detection framework is its role as a modular, plug-in defense. Rather than replacing robust training methods, our detector can be integrated with existing robust models, such as Adversarially Trained Classifiers (ATCs), to further enhance their performance. The value of a detector in this context is measured by the \textbf{robustness improvement} it provides to the overall system.

\subsubsection{System Robustness Identity and Independence Approximation}
Let $K=\{C(x_{\mathrm{adv}})=y\}$ and $Q=\{D(x_{\mathrm{adv}})=1\}$. The exact identity is
\begin{equation}
RA_{sys}=\Pr(K\cup Q)=RA_{cls}+\Pr(Q\cap K^c).
\end{equation}
Consequently $RA_{sys}\geq RA_{cls}$, with strict improvement iff the detector flags a nonzero fraction of adversarial inputs that the classifier misclassifies. In terms of marginal $TPR=\Pr(Q)$ alone,
\begin{equation}
\max(RA_{cls},TPR)\leq RA_{sys}\leq\min(1,RA_{cls}+TPR).
\end{equation}
If $Q$ and $K$ are independent, this identity reduces to $TPR+(1-TPR)RA_{cls}$. \autoref{tab:atc_gains_empirical} reports this independence approximation alongside the exact measured $RA_{sys}$.

\subsubsection{Empirical Validation}
We apply the detector to several ATCs with varying baseline robustness. The results in \autoref{tab:atc_gains_empirical} show two points:
\begin{itemize}
    \item The measured $RA_{sys}$ shows a substantial improvement over the initial $RA_{cls}$ in all cases. For instance, a classifier with a baseline robustness of 55.88\% achieves a final robustness of 83.08\% when paired with our detector.
    \item The absolute differences between measured $RA_{sys}$ and the independence approximation are 3.41, 6.23, and 5.02 percentage points. These differences reflect dependence between classifier correctness and detector firing.
\end{itemize}

\begin{table}[h]
\centering
\caption{Independence approximation and measured System Robust Accuracy ($RA_{sys}$) (\%) when applying our detector to Adversarially Trained Classifiers with varying baseline robustness ($RA_{cls}$) under the reported adaptive attack. The final column is measured $RA_{sys}-RA_{cls}$ in percentage points.}
\label{tab:atc_gains_empirical}
\resizebox{\textwidth}{!}{
\begin{tabular}{ccccc}
\toprule
\textbf{$RA_{cls}$ (\%)} & \textbf{Detector TPR (\%)} & \textbf{Independence approx. (\%)} & \textbf{Measured $RA_{sys}$ (\%)} & \textbf{Increase over $RA_{cls}$ (points)} \\
\midrule
6.84 & 72.13 & 74.03 & 77.44 & +70.60 \\
12.71 & 63.74 & 68.34 & 74.57 & +61.86 \\
55.88 & 73.03 & 88.10 & 83.08 & +27.20 \\
\bottomrule
\end{tabular}
}
\end{table}

\subsection{System Accuracy Lower Bounds}
\textbf{General Lower Bound.} When the attack probability $p$ is unknown, we can establish a general lower bound on system accuracy.
\begin{theorem}
Let $p, CA_{sys}, RA_{sys} \in[0, 1]$. Then $(1-p)CA_{sys} + p \cdot RA_{sys} \geq CA_{sys} \cdot RA_{sys}$.
\end{theorem}
\begin{proof}
Let $f(p) = (1-p)CA_{sys} + p \cdot RA_{sys} - CA_{sys} \cdot RA_{sys}$. As a linear function of $p$ over the interval $[0, 1]$, its minimum must occur at an endpoint. At $p=0$, $f(0) = CA_{sys}(1-RA_{sys}) \ge 0$. At $p=1$, $f(1) = RA_{sys}(1-CA_{sys}) \ge 0$. Since the function is non-negative at both endpoints, the inequality holds for all $p \in [0, 1]$.
\end{proof}

\textbf{Bound with known maximum attack probability.} For practical scenarios where we can assume an upper bound on the attack probability ($p \le p'$), we can derive a tighter lower bound.
\begin{theorem}
Let $p \in [0, p']$ and $CA_{sys}, RA_{sys}, p' \in [0, 1]$. Then the system accuracy is lower-bounded by:
\begin{equation}
(1-p)CA_{sys} + p \cdot RA_{sys} \geq CA_{sys} \cdot RA_{sys} + \max(0, (1 - p')(CA_{sys}-RA_{sys}))
\end{equation}
\end{theorem}
\begin{proof}
Write $C=CA_{sys}$ and $R=RA_{sys}$. If $C<R$, the added maximum is zero and the result is the general bound above. If $C\geq R$, $(1-p)C+pR$ is decreasing in $p$, so for $p\leq p'$ it is at least $(1-p')C+p'R$. Moreover,
\begin{equation}
(1-p')C+p'R-\{CR+(1-p')(C-R)\}=R(1-C)\geq0.
\end{equation}
Combining the two cases proves the result.
\end{proof}

\subsection{Empirical Validation and Operating-Point Selection}
The operating point follows from the assumed attack-probability range and the measured $CA_{sys}$ and $RA_{sys}$. For the FGSM results in \autoref{tab:lower_bound_analysis}, maximizing $CA_{sys}RA_{sys}$ selects FPR 10\%. When the assumed maximum attack probability is instead $p'\leq1\%$, maximizing the corresponding bound selects FPR 1\%.

\begin{table}[h]
\centering
\caption{System performance and lower bounds (\%) at different FPRs for the FGSM attack ($\epsilon = 8/255$). The general-bound column is $CA_{sys}RA_{sys}$.}
\label{tab:lower_bound_analysis}
\resizebox{\textwidth}{!}{
\begin{tabular}{r|ccc|c|cccccccc|cccccccc}
\toprule
\textbf{FPR} & \textbf{TPR} & \textbf{$CA_{sys}$} & \textbf{$RA_{sys}$} & \textbf{General Bound} & \multicolumn{8}{c|}{\textbf{$\boldsymbol{A_{sys}}$ for attack prob. $\boldsymbol{p}$}} & \multicolumn{8}{c}{\textbf{Bound for max attack prob. $\boldsymbol{p'}$}} \\
 & & & & & \textbf{0.1\%} & \textbf{1\%} & \textbf{5\%} & \textbf{10\%} & \textbf{50\%} & \textbf{90\%} & \textbf{95\%} & \textbf{99\%} & \textbf{0.1\%} & \textbf{1\%} & \textbf{5\%} & \textbf{10\%} & \textbf{50\%} & \textbf{90\%} & \textbf{95\%} & \textbf{99\%} \\
\midrule
1 & 57.67 & \textbf{91.77} & 71.43 & 65.55 & \textbf{91.75} & \textbf{91.57} & \textbf{90.75} & \textbf{89.74} & 81.60 & 73.46 & 72.45 & 71.63 & \textbf{85.87} & \textbf{85.69} & \textbf{84.87} & \textbf{83.86} & 75.72 & 67.59 & 66.57 & 65.75 \\
5 & 86.79 & 88.53 & 91.54 & 81.04 & 88.53 & 88.56 & 88.68 & 88.83 & 90.04 & 91.24 & 91.39 & 91.51 & 81.04 & 81.04 & 81.04 & 81.04 & 81.04 & 81.04 & 81.04 & 81.04 \\
10 & 95.42 & 84.60 & 97.26 & \textbf{82.28} & 84.61 & 84.73 & 85.23 & 85.87 & \textbf{90.93} & 95.99 & 96.63 & 97.13 & 82.28 & 82.28 & 82.28 & 82.28 & \textbf{82.28} & \textbf{82.28} & \textbf{82.28} & \textbf{82.28} \\
15 & 97.75 & 80.34 & 98.64 & 79.25 & 80.36 & 80.52 & 81.25 & 82.17 & 89.49 & 96.81 & 97.73 & 98.46 & 79.25 & 79.25 & 79.25 & 79.25 & 79.25 & 79.25 & 79.25 & 79.25 \\
20 & 99.00 & 76.07 & 99.42 & 75.63 & 76.09 & 76.30 & 77.24 & 78.40 & 87.74 & \textbf{97.08} & \textbf{98.25} & 99.19 & 75.63 & 75.63 & 75.63 & 75.63 & 75.63 & 75.63 & 75.63 & 75.63 \\
25 & 99.34 & 71.66 & 99.63 & 71.39 & 71.69 & 71.94 & 73.06 & 74.46 & 85.64 & 96.83 & 98.23 & 99.35 & 71.39 & 71.39 & 71.39 & 71.39 & 71.39 & 71.39 & 71.39 & 71.39 \\
30 & \textbf{99.63} & 67.21 & \textbf{99.80} & 67.08 & 67.24 & 67.54 & 68.84 & 70.47 & 83.51 & 96.54 & 98.17 & \textbf{99.47} & 67.08 & 67.08 & 67.08 & 67.08 & 67.08 & 67.08 & 67.08 & 67.08 \\
\bottomrule
\end{tabular}
}
\end{table}

\section{Robustness to Benign Noise}
\label{app:benign_noise}

We evaluate prediction-preserving random noise of varying magnitude on the held-out test set, retaining only instances that do not change the classifier's original prediction. We then apply thresholds calibrated on clean, unperturbed data; a broader benign-shift analysis follows in \autoref{app:calibration_failure}.

In \autoref{tab:benign_noise_robustness}, FPR remains close to its clean operating point at $\epsilon=4/255$ and $8/255$, but rises to 17.51\% at $\epsilon=32/255$ for a threshold calibrated at clean FPR 5\%. The result demonstrates low-noise stability in this experiment and also shows that sufficiently large benign shift can be over-flagged, consistent with the deployment-recalibration analysis in \autoref{app:calibration_failure}.

\begin{table}[h]
\centering
\caption{Measured False Positive Rate (FPR) (\%) on Benign Data with Random Noise. This table shows the new FPR when applying thresholds that were originally calibrated to give 1\%, 5\%, 10\%, etc., FPR on clean data. The test is repeated for different magnitudes ($\epsilon$) of benign random noise.}
\label{tab:benign_noise_robustness}
\resizebox{\textwidth}{!}{
\begin{tabular}{lccccccccccc}
\toprule
$\boldsymbol{\epsilon}$ & \textbf{FPR@1\%} & \textbf{FPR@5\%} & \textbf{FPR@10\%} & \textbf{FPR@15\%} & \textbf{FPR@20\%} & \textbf{FPR@25\%} & \textbf{FPR@30\%} & \textbf{FPR@35\%} & \textbf{FPR@40\%} & \textbf{FPR@45\%} & \textbf{FPR@50\%} \\
\midrule
\textbf{4/255} & 0.93 & 4.77 & 9.27 & 13.60 & 18.51 & 23.03 & 27.61 & 32.01 & 36.27 & 41.20 & 46.26 \\
\textbf{8/255} & 1.05 & 4.46 & 8.04 & 11.79 & 15.65 & 19.60 & 23.52 & 27.65 & 32.07 & 36.61 & 41.15 \\
\textbf{32/255} & 12.38 & 17.51 & 22.16 & 26.71 & 31.39 & 35.99 & 40.34 & 44.72 & 49.16 & 53.54 & 57.85 \\
\bottomrule
\end{tabular}
}
\end{table}

\section{Deployment Calibration and Failure-Case Analysis}
\label{app:calibration_failure}

\subsection{Finite-Sample Marginal FPR Control}

\begin{proposition}[Exchangeable benign calibration]\label{prop:conformal_fpr}
Let the complete score function $S(\cdot)$ be fixed independently of $n$ benign calibration samples. If calibration scores $S_1,\ldots,S_n$ and a future benign score $S_{n+1}$ are exchangeable, set $k=\lceil(n+1)(1-\alpha)\rceil$, $\tau_\alpha=S_{(k)}$, and flag when $S_{n+1}>\tau_\alpha$ (using $\tau_\alpha=+\infty$ if $k>n$). Then
\begin{equation}
\Pr\{S_{n+1}>\tau_\alpha\}\leq\alpha.
\end{equation}
\end{proposition}

\begin{proof}
Under exchangeability, the rank of a future score among the $n+1$ scores is uniform when scores are distinct. Rejection can occur only for one of the $n+1-k$ ranks above $k$, so its probability is $(n+1-k)/(n+1)\leq\alpha$. With ties, the strict rejection rule is conservative.
\end{proof}

The proposition provides marginal finite-sample control over the random calibration set and future benign input, with FPR resolution $1/(n+1)$. It requires exchangeability between calibration and future benign scores; shifted deployment populations therefore require representative recalibration data.

\subsection{Calibration of the Learned RLT Detector}

We test the complete learned RLT detector on CIFAR-10/ResNet-110. We use 40,000 benign training images to fit the recovery modules and LT transformations, and a disjoint 5,000-image benign training subset to fix the empirical CDFs. We then randomly split the official 10,000-image test population into 5,000 images for threshold calibration and 5,000 for evaluation. Thus the score function is fixed before threshold calibration, and the calibration and evaluation samples are disjoint and exchangeable under random sampling from the test population.

\begin{table}[h]
\centering
\caption{Calibration-size sensitivity of the learned RLT score at nominal FPR 5\%. Entries are held-out FPR mean $\pm$ standard deviation and 95th percentile across 100 calibration resamples; all are evaluated on the same disjoint 5,000-image evaluation split.}
\label{tab:actual_rlt_calibration_size}
\begingroup
\resizebox{0.6\textwidth}{!}{
\begin{tabular}{rccc}
\toprule
\textbf{Calibration size} & \textbf{Held-out FPR (\%)} & \textbf{95th percentile (\%)} & \textbf{Resolution (\%)} \\
\midrule
100   & $5.25\pm2.00$ & 8.32 & 0.990 \\
250   & $5.04\pm1.17$ & 7.04 & 0.398 \\
500   & $5.50\pm0.89$ & 7.12 & 0.200 \\
1,000 & $5.31\pm0.59$ & 6.20 & 0.100 \\
2,500 & $5.51\pm0.25$ & 5.76 & 0.040 \\
5,000 & 5.58            & 5.58 & 0.020 \\
\bottomrule
\end{tabular}
}
\endgroup
\end{table}

With all 5,000 calibration images, the held-out FPR is 5.58\% (279/5,000; 95\% Wilson interval $[4.98,6.25]\%$), statistically compatible with the nominal 5\% rate. The resampling variation decreases from 2.00 percentage points at $n=100$ to 0.25 at $n=2{,}500$, while the order-statistic resolution decreases deterministically as $1/(n+1)$. These results distinguish finite-sample variability from the marginal guarantee in \autoref{prop:conformal_fpr}.

\begin{table}[h]
\centering
\caption{Benign distribution shift for the learned RLT detector at nominal FPR 5\%. ``Source threshold'' applies the clean-calibrated threshold to shifted evaluation inputs. ``Target recalibrated'' computes a new threshold from the corresponding shifted calibration split while keeping the learned detector and empirical CDFs fixed.}
\label{tab:actual_rlt_shift_recalibration}
\begingroup
\resizebox{0.70\textwidth}{!}{
\begin{tabular}{lcc}
\toprule
\textbf{Benign deployment domain} & \textbf{Source-threshold FPR (\%)} & \textbf{Target-recalibrated FPR (\%)} \\
\midrule
Brightness      & 5.22  & 5.36 \\
Contrast        & 4.68  & 4.50 \\
Gaussian blur   & 6.46  & 5.04 \\
Pixelation      & 78.78 & 4.88 \\
Gaussian noise  & 35.00 & 5.22 \\
\bottomrule
\end{tabular}
}
\endgroup
\end{table}

Mild brightness and contrast changes remain close to nominal under the source threshold, whereas severe pixelation and Gaussian noise are over-flagged. Target-domain threshold calibration restores all five evaluated FPRs to $4.50$--$5.36\%$. Representative target-domain benign data are therefore required for deployment, and larger model or pipeline changes may require refitting the CDFs or auxiliary modules before a separate final-threshold calibration.

\subsection{Calibration Sensitivity in the Expanded AFLS--LT Analysis}

The preceding study evaluates calibration of the complete learned RLT detector. We next examine sensitivity across 12 dataset/model combinations using two quantities from the expanded AFLS experiment: the largest $A_k$ computed from each clean and perturbed input, and the corresponding LT score. Because $A_k$ uses the clean counterpart, this paired-input analysis specifically examines how calibration-set size and benign-perturbation coverage affect CDF normalization and threshold calibration across a broader set of models.

For each dataset/model pair, we apply the clipped CDF-to-normal transformation from \autoref{eq:rlt} to the largest $A_k$ and LT, sum their squared normalized values, and fit a nominal 5\% FPR threshold on a randomly sampled benign subset. Evaluation uses disjoint benign samples and inputs generated by FGSM, PGD, CW, and AutoAttack. The learned-RLT study above provides the corresponding calibration analysis for the test-time detector.

\begin{table}[h]
\centering
\caption{Calibration-set-size sensitivity of the auxiliary AFLS--LT analysis, averaged over 12 dataset/model combinations and 20 random resamples. A nominal 5\% threshold is fitted on the sampled benign calibration set and evaluated on held-out benign and attack-generated samples.}
\label{tab:calibration_size_sensitivity}
\begingroup
\resizebox{0.70\textwidth}{!}{
\begin{tabular}{rcc}
\toprule
\textbf{Benign calibration samples} & \textbf{Held-out FPR (\%)} & \textbf{Attack detection rate (\%)} \\
\midrule
100 & $6.13_{\pm2.63}$ & $83.92_{\pm13.65}$ \\
250 & $5.50_{\pm1.61}$ & $87.44_{\pm10.02}$ \\
500 & $5.03_{\pm1.20}$ & $87.20_{\pm9.48}$ \\
1,000 & $4.92_{\pm0.87}$ & $87.77_{\pm8.96}$ \\
2,500 & $4.96_{\pm0.68}$ & $88.50_{\pm8.55}$ \\
5,000 & $4.93_{\pm0.59}$ & $88.48_{\pm8.52}$ \\
10,000 & $4.95_{\pm0.53}$ & $88.69_{\pm8.52}$ \\
\bottomrule
\end{tabular}
}
\endgroup
\end{table}

\autoref{tab:calibration_size_sensitivity} shows that the auxiliary combined score is close to nominal FPR with 500--1,000 representative benign samples and that larger sets mainly reduce variance. Deliberately representing benign variation with only one transformation can over-flag other shifts (\autoref{tab:calibration_distribution_coverage}); pooling crop, horizontal flip, and noise restores near-nominal FPR for those three perturbations. This sensitivity pattern agrees with the learned RLT result that threshold calibration should represent the intended benign population.

\begin{table}[h]
\centering
\caption{Effect of benign-perturbation coverage in the expanded AFLS--LT analysis. Each row uses 5,000 calibration samples and a nominal 5\% threshold. Calibration on one transformation intentionally under-represents benign variation; pooled calibration covers the three evaluated benign perturbations.}
\label{tab:calibration_distribution_coverage}
\begingroup
\resizebox{0.85\textwidth}{!}{
\begin{tabular}{lcccc}
\toprule
\textbf{Calibration pool} & \textbf{FPR on crop (\%)} & \textbf{FPR on hflip (\%)} & \textbf{FPR on noise (\%)} & \textbf{Attack detection rate (\%)} \\
\midrule
Crop only & 5.00 & 26.61 & 22.31 & 96.12 \\
Horizontal flip only & 23.68 & 5.00 & 23.27 & 97.76 \\
Noise only & 65.60 & 54.25 & 5.01 & 79.83 \\
Pooled benign shifts & 5.61 & 4.26 & 4.80 & 88.41 \\
\bottomrule
\end{tabular}
}
\endgroup
\end{table}

We also separate attack-generated inputs according to whether the largest $A_k$ and LT exceed their respective benign reference bounds. In \autoref{tab:rt_lt_failure_modes}, ``$A_k$ only'' denotes an $A_k$ exceedance without an LT exceedance, ``LT only'' the converse, ``Both'' an exceedance by both criteria, and ``Neither'' an exceedance by neither. LT alone identifies 9.56\% of CW and 11.37\% of AutoAttack inputs for which every $A_k$ remains within its benign bound. This decomposition analyzes the direct AFLS statistic; the learned RT score is evaluated separately in the main detector experiments.

\begin{table}[h]
\centering
\caption{Percentage of attack-generated samples for which the largest $A_k$, LT, both statistics, or neither statistic exceed their respective benign reference bounds, averaged over 12 dataset/model combinations.}
\label{tab:rt_lt_failure_modes}
\begingroup
\resizebox{0.55\textwidth}{!}{
\begin{tabular}{lcccc}
\toprule
\textbf{Attack} & \textbf{$A_k$ only (\%)} & \textbf{LT only (\%)} & \textbf{Both (\%)} & \textbf{Neither (\%)} \\
\midrule
FGSM & 93.91 & 0.01 & 0.00 & 6.08 \\
PGD & 99.94 & 0.00 & 0.00 & 0.06 \\
CW & 75.99 & 9.56 & 0.44 & 14.01 \\
AutoAttack & 79.75 & 11.37 & 0.00 & 8.88 \\
\midrule
Average & 87.40 & 5.23 & 0.11 & 7.26 \\
\bottomrule
\end{tabular}
}
\endgroup
\end{table}

\section{Attack Configuration Summary}
\label{app:attack_configurations}

We summarize the attack configurations used in the main experiments and appendices in \autoref{tab:attack_configurations}; settings inherited from a baseline protocol or implementation are identified as such. For the standard comparison tables, we follow the perturbation budgets used by the corresponding baselines: FGSM uses $\epsilon=0.05$, PGD uses $\epsilon=0.02$, and the remaining standard attacks use $\epsilon=8/255$ unless otherwise specified. For iterative attacks, we use 50 steps with absolute step size $0.002$ when the attack implementation does not determine the step size internally. Adaptive attacks are evaluated up to $\epsilon=8/255$.

\begin{table*}[t]
\centering
\caption{\textbf{Attack configurations used in the standard and adaptive evaluations.}
``Default'' denotes the public/default implementation recommended by the original attack
or by the baseline protocol used for comparison.}
\label{tab:attack_configurations}
\begingroup
\scriptsize
\setlength{\tabcolsep}{3pt}
\renewcommand{\arraystretch}{1.15}

\newcolumntype{L}[1]{>{\RaggedRight\arraybackslash}p{#1}}
\newcolumntype{Y}{>{\RaggedRight\arraybackslash}X}

\begin{tabularx}{\textwidth}{@{}
L{0.10\textwidth}
L{0.13\textwidth}
Y
L{0.10\textwidth}
L{0.12\textwidth}
L{0.14\textwidth}
L{0.15\textwidth}
@{}}
\toprule
\textbf{Attack}
&
\textbf{Threat model}
&
\textbf{Objective / loss}
&
\textbf{Budget}
&
\textbf{Steps / queries}
&
\textbf{Step size / search}
&
\textbf{Initialization / restarts}
\\
\midrule

FGSM
&
Standard white-box
&
Cross-entropy\newline untargeted
&
$\epsilon=0.05$ in baseline comparison;\newline
$8/255$ in expanded validation
&
1
&
Single sign step
&
Deterministic from clean input
\\

PGD
&
Standard white-box
&
Cross-entropy\newline untargeted
&
$\epsilon=0.02$ in baseline comparison;\newline
$8/255$ in expanded validation
&
50
&
$0.002$ absolute step size
&
Random start;\newline
conventional or default restarts
\\

CW
&
Standard white-box
&
Carlini--Wagner\newline margin loss
&
$8/255$
&
50
&
Default or conventional binary search\newline
or $c$ search when available
&
Default initialization/restarts
\\

AutoAttack
&
Standard ensemble
&
APGD-CE,\newline
APGD-DLR,\newline
FAB,\newline
Square ensemble
&
$8/255$
&
Default standard version
&
Default internal schedule
&
Default deterministic ensemble
\\

Square
&
Query-based\newline black-box
&
Score-based square perturbation search
&
$8/255$
&
5,000 queries
&
$p_{\mathrm{init}}=0.8$
&
1 restart
\\

Orthogonal-PGD
&
Adaptive white-box\newline detector attack
&
Orthogonalized classification\newline
and detector-evasion directions
&
$0.01$ and $8/255$
&
Baseline protocol of~\cite{bryniarski2021evading,he2024beyond}
&
Orthogonal projection removes\newline
scalar-loss balancing sensitivity
&
Protocol-matched initialization/restarts
\\

End-to-end PGD + BPDA
&
Adaptive white-box\newline detector attack
&
$\min_{\|\delta\|_\infty\le\epsilon}
\{-\mathcal{L}_{cls}(x+\delta,y)+\lambda RLT(x+\delta)\}$
&
$8/255$
&
50
&
$0.002$ absolute step size
&
Random start;\newline
5 repeated runs;\newline
$\lambda\in\{1.00,0.50,0.25\}$
\\

End-to-end PGD + BPDA + ATC
&
Adaptive white-box\newline detector attack with ATC
&
Same as above with adversarially trained classifier
&
$8/255$
&
50
&
$0.002$ absolute step size
&
Random start;\newline
5 repeated runs;\newline
$\lambda\in\{1.00,0.50,0.25\}$
\\

SimBA
&
Gradient-masking\newline check
&
Gradient-free logit/probability\newline query attack; detector score not queried
&
$8/255$
&
1,000 queries
&
Coordinate-query updates
&
Default query initialization
\\

$L_2$-PGD
&
Norm-transfer appendix
&
Cross-entropy\newline untargeted
&
$\epsilon=0.6$, the first tested budget with base $RA=0\%$
&
50
&
Conventional or default\newline $L_2$ PGD schedule
&
Random start
\\

\bottomrule
\end{tabularx}
\endgroup
\end{table*}

\clearpage
\section{Additional Empirical Evaluation of the A Few Large Shifts Hypothesis}
\label{app:additional_afls_validation}

We evaluate the \textbf{A Few Large Shifts} hypothesis on CIFAR-10, CIFAR-100, and ImageNet with ResNet-18, ResNet-34, VGG-11, and VGG-13. For each clean input $x$ and corresponding version $x'$ produced by a benign transformation or attack, we compute the dimension-normalized representation change $d_k$ and its regularized growth $A_k$ as defined in Section~\ref{sec:afls_hypothesis}, using $\varepsilon_A=10^{-4}$ and $\varepsilon_C=10^{-8}$. For each pair of consecutive stages, the reference bound $b_k$ is the pooled mean plus three standard deviations over random crop, horizontal flip, and random noise; the LT reference bound is constructed analogously. We report the probability $\Pr(V=1)$ that at least one $A_k$ exceeds its bound and the mean concentration $C_1$ at the largest $A_k$ over all generated samples.

\begin{figure}[h]
    \centering
    \begin{overpic}[width=1.0\linewidth]{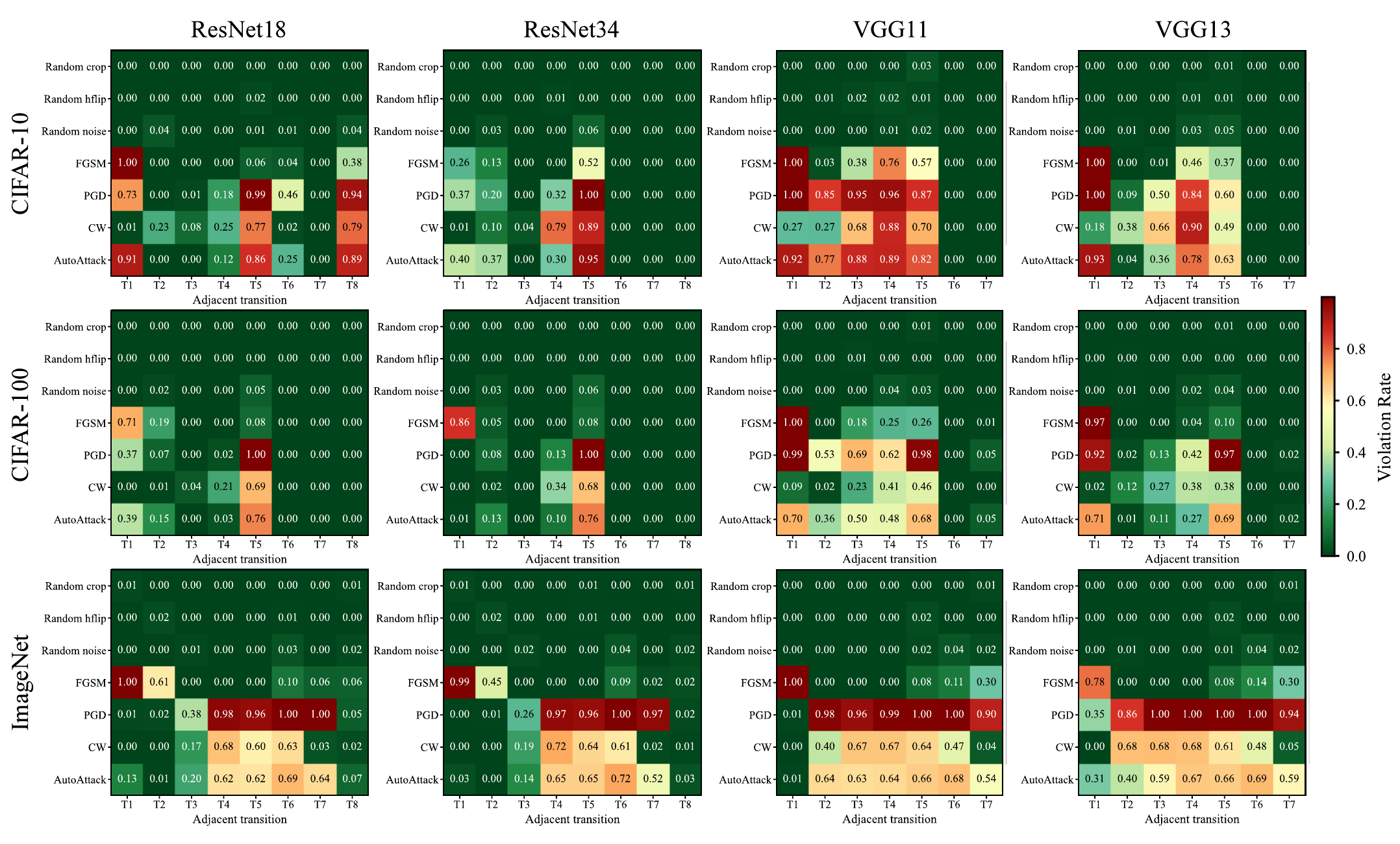}
        \put(97.7,20.5){\rotatebox{90}{\colorbox{white}{\makebox[1.45in][c]{\color{revisionthree}\sffamily\scriptsize Bound Exceedance Rate}}}}
    \end{overpic}
    \caption{\textbf{Growth of normalized representation changes across consecutive network stages.} Each panel reports, for one dataset/model combination, the frequency with which $A_k$ exceeds its benign reference bound under random crop, horizontal flip, random noise, FGSM, PGD, CW, and AutoAttack. Benign exceedance rates are generally low, while attacks frequently exceed at least one bound. Multiple values can exceed their bounds; concentration at the largest value is reported separately in \autoref{tab:compact_afls_summary}.}
    \label{fig:layerwise_lipschitz_violation}
\end{figure}

\begingroup
\color{blue}
\begin{figure}[h]
    \centering
    \includegraphics[width=1.0\linewidth]{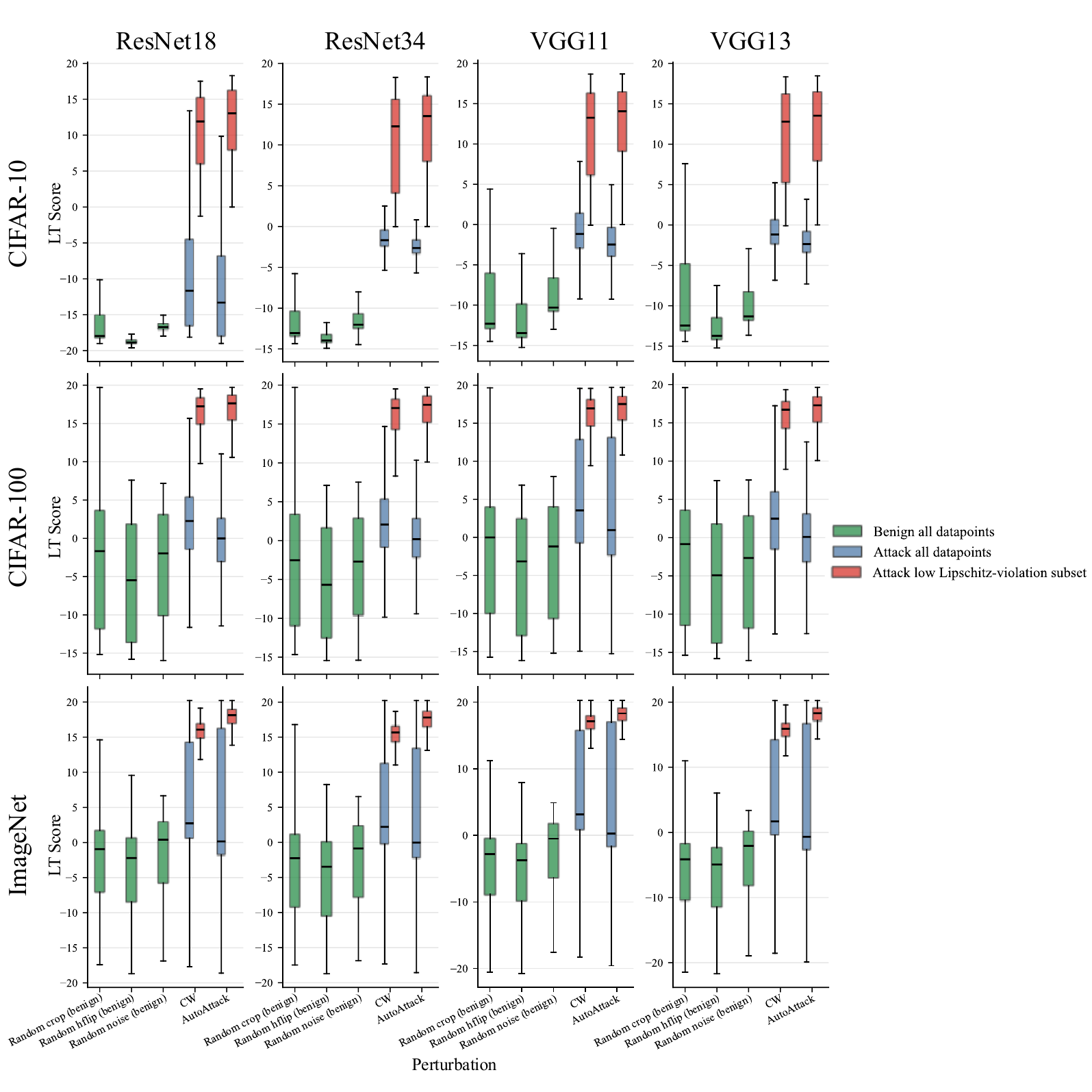}
    \caption{\textbf{LT distributions conditional on no AFLS bound exceedance.} Each panel compares LT scores for benign perturbations with scores for CW and AutoAttack inputs. For the subset in which every $A_k$ remains within its benign reference bound, the attack-generated LT scores shift upward relative to the corresponding distribution over all attack-generated samples, showing that the two quantities capture complementary behavior.}
    \label{fig:lt_distributions}
\end{figure}
\endgroup

\begin{table*}[h]
\centering
\caption{\textbf{Compact AFLS evaluation summary.} ``Bound exceedance'' is $100\Pr(V=1)$, the percentage of samples with at least one $A_k$ above its benign reference bound. ``Concentration at largest $A_k$'' is $100\,\mathbb E[C_1]$, the mean percentage of total amplification accounted for by the largest value. ``LT only'' is the percentage of CW/AutoAttack samples for which every $A_k$ remains within its bound but LT exceeds its benign bound. The final column gives the consecutive stages at which $A_k$ is largest most frequently.}
\label{tab:compact_afls_summary}
\scriptsize
\resizebox{\textwidth}{!}{%
\begin{tabular}{llccccccl}
\toprule
\multirow{2}{*}{\textbf{Dataset}} & \multirow{2}{*}{\textbf{Model}} &
\multicolumn{2}{c}{\textbf{Benign bound exceedance (\%)}} &
\multicolumn{2}{c}{\textbf{Attack bound exceedance (\%)}} &
\textbf{Attack concentration} &
\textbf{CW/AA} &
\multirow{2}{*}{\textbf{Most frequent location of largest $A_k$}} \\
\cmidrule(lr){3-4}\cmidrule(lr){5-6}
& & \textbf{Mean} & \textbf{Max} & \textbf{Mean} & \textbf{Min} & \textbf{at largest $A_k$ (\%)} & \textbf{LT only (\%)} & \\
\midrule
\multirow{4}{*}{CIFAR-10}
& ResNet-18 & 3.64 & 8.82 & 97.50 & 94.40 & 89.91 & 3.68 & embedding$\rightarrow$logit \\
& ResNet-34 & 3.27 & 8.56 & 89.78 & 69.31 & 75.84 & 3.72 & embedding$\rightarrow$logit \\
& VGG-11    & 4.13 & 5.00 & 95.82 & 91.15 & 67.01 & 6.10 & embedding$\rightarrow$logit \\
& VGG-13    & 4.35 & 8.73 & 96.79 & 93.25 & 62.71 & 4.63 & embedding$\rightarrow$logit \\
\midrule
\multirow{4}{*}{CIFAR-100}
& ResNet-18 & 2.36 & 6.96 & 83.46 & 72.36 & 59.80 & 0.70 & res3$\rightarrow$res4 \\
& ResNet-34 & 3.14 & 9.00 & 84.14 & 70.92 & 50.83 & 7.90 & res3$\rightarrow$res4 \\
& VGG-11    & 2.86 & 6.29 & 82.26 & 58.41 & 79.61 & 0.00 & vgg4$\rightarrow$vgg5 \\
& VGG-13    & 2.74 & 7.12 & 83.94 & 63.46 & 77.57 & 0.07 & vgg4$\rightarrow$vgg5 \\
\midrule
\multirow{4}{*}{ImageNet}
& ResNet-18 & 4.19 & 6.44 & 84.56 & 68.99 & 75.69 & 22.30 & res3$\rightarrow$res4 \\
& ResNet-34 & 4.58 & 7.54 & 86.22 & 72.52 & 71.36 & 19.23 & res3$\rightarrow$res4 \\
& VGG-11    & 3.74 & 7.55 & 84.09 & 68.16 & 53.23 & 29.03 & input$\rightarrow$vgg1 \\
& VGG-13    & 3.86 & 7.61 & 81.58 & 69.14 & 78.44 & 28.23 & input$\rightarrow$vgg1 \\
\midrule
\multicolumn{2}{l}{\textbf{Overall mean}} & \textbf{3.57} & -- & \textbf{87.51} & -- & \textbf{70.17} & \textbf{10.47} & -- \\
\bottomrule
\end{tabular}
}
\end{table*}

\begin{table}[h]
\centering
\caption{\textbf{Consecutive stages represented by $T_i$ in \autoref{fig:layerwise_lipschitz_violation}.} The feature list differs by model family.}
\label{tab:transition_label_legend}
\scriptsize
\resizebox{\textwidth}{!}{
\begin{tabular}{ll}
\toprule
\textbf{Model family} & \textbf{Consecutive stages} \\
\midrule
CIFAR ResNet-18/34 &
$T_1$: input$\rightarrow$stem; $T_2$: stem$\rightarrow$res1; $T_3$: res1$\rightarrow$res2; $T_4$: res2$\rightarrow$res3; $T_5$: res3$\rightarrow$res4; $T_6$: res4$\rightarrow$pool; $T_7$: pool$\rightarrow$embedding; $T_8$: embedding$\rightarrow$logit \\
ImageNet ResNet-18/34 &
$T_1$: input$\rightarrow$stem; $T_2$: stem$\rightarrow$maxpool; $T_3$: maxpool$\rightarrow$res1; $T_4$: res1$\rightarrow$res2; $T_5$: res2$\rightarrow$res3; $T_6$: res3$\rightarrow$res4; $T_7$: res4$\rightarrow$embedding; $T_8$: embedding$\rightarrow$logit \\
CIFAR/ImageNet VGG-11/13 &
$T_1$: input$\rightarrow$vgg1; $T_2$: vgg1$\rightarrow$vgg2; $T_3$: vgg2$\rightarrow$vgg3; $T_4$: vgg3$\rightarrow$vgg4; $T_5$: vgg4$\rightarrow$vgg5; $T_6$: vgg5$\rightarrow$embedding; $T_7$: embedding$\rightarrow$logit \\
\bottomrule
\end{tabular}
}
\end{table}

\clearpage
\section{Detailed Algorithm}
\label{app:detailalgorithm}

\autoref{alg:layerwise_detection} presents the detailed pseudocode for RT, LT, and the combined RLT measure.

\begin{algorithm}
\caption{Layer-wise Adversarial Detection Measures via RT, LT, and RLT}
\label{alg:layerwise_detection}
\begin{algorithmic}[1]
\Require 
    $f = f_{logit} \circ f_L \circ \dots \circ f_1$: Target network \\
    $\{R^{(L \rightarrow k)}\}_{k=k_{RT}}^{L-1}$: Trained inverse regressors \\
    $\{W^{(g)}\}_{g=1}^G$: Learned augmentation matrices \\
    $\widetilde{\mathcal{F}}_{RT}, \widetilde{\mathcal{F}}_{LT}$: Clipped mid-rank empirical CDFs of RT and LT (from benign data) \\
    $\Phi^{-1}$: Standard normal quantile function \\
    $x$: Test input

\Statex
\Function{RT}{$x$}
    \For{$k = k_{RT}$ to $L-1$}
        \State $e_k \gets \| z_k(x) - R^{(L \rightarrow k)}(z_L(x)) \|_2^2$
    \EndFor
    \State $\bm{e}_{RT} \gets (e_{k_{RT}}, \dots, e_{L-1})$
    \State \Return $(\log (L-k_{RT}) - \mathcal{H}(\sigma(\bm{e}_{RT}))) \cdot \log \left( \frac{1}{L-k_{RT}} \sum_{k=k_{RT}}^{L-1} e_k \right)$
\EndFunction

\Statex
\Function{LT}{$x$}
    \For{$g = 1$ to $G$}
        \State $\Delta z^{(g)} \gets \frac{1}{L - k_{LT} + 1} \sum_{i=k_{LT}}^{L} \| z_i(x) - z_i(W^{(g)}x) \|_2^2$
        \State $\hat{y} \gets \arg\max_c \sigma_c(\ell(x))$
        \State $\Delta \ell^{(g)} \gets \| \mathbf{o}_{\hat{y}} - \sigma(\ell(W^{(g)}x)) \|_2^2$
        \State $s^{(g)} \gets \log \left( \mathcal{H}(\sigma(\ell(x))) \cdot \Delta \ell^{(g)} + \varepsilon \right) - \log \left(\Delta z^{(g)}+\varepsilon\right)$
    \EndFor
    \State \Return $\frac{1}{G} \sum_{g=1}^G s^{(g)}$
\EndFunction

\Statex
\Function{RLT}{$x$}
    \State $r \gets$ \Call{RT}{$x$}, \quad $l \gets$ \Call{LT}{$x$}
    \State $r_{norm} \gets \Phi^{-1}(\widetilde{\mathcal{F}}_{RT}(r))$
    \State $l_{norm} \gets \Phi^{-1}(\widetilde{\mathcal{F}}_{LT}(l))$
    \State \Return $r_{norm}^2 + l_{norm}^2$
\EndFunction
\end{algorithmic}
\end{algorithm}

\end{document}